\documentclass{article}

 \usepackage[preprint]{neurips_2026}

\usepackage[utf8]{inputenc} 
\usepackage[T1]{fontenc}    
\usepackage{hyperref}       
\usepackage{url}            
\usepackage{booktabs}       
\usepackage{graphicx}
\usepackage{amsfonts}       
\usepackage{amsmath}
\usepackage{amssymb, centernot}
\usepackage{mathtools}
\usepackage{amsthm}
\usepackage{caption}
\usepackage{enumitem}

\usepackage{nicefrac}       
\usepackage{microtype}      
\usepackage{xcolor}         
\usepackage{algorithm}
\usepackage{algpseudocode}
\newcommand{\nll}{\centernot{\ll}}
\newcommand{\norm}[1]{\left\lVert#1\right\rVert}
\newtheorem{conjecture}{Conjecture}

\usepackage{subcaption}

\theoremstyle{plain}
\newtheorem{theorem}{Theorem}[section]

\theoremstyle{definition}

\newtheorem{assumption}[theorem]{Assumption}
\theoremstyle{remark}

\title{Implicit Hypothesis Testing and Divergence Preservation in Neural Network Representations}

\author{%
  Kadircan Aksoy \\
  German Aerospace Center, Institute for Space Research\\
  Technical University of Berlin\\
  Berlin, Germany \\
  \texttt{kadircan.aksoy@dlr.de} \\
\And
  Protim Bhattacharjee \\
  German Aerospace Center, Institute for Space Research\\
  Berlin, Germany \\
  \texttt{protim.bhattacharjee@dlr.de} \\
\And
  Peter Jung \\
  German Aerospace Center, Institute for Space Research\\
  Technical University of Berlin\\
  Berlin, Germany \\
  \texttt{peter.jung@dlr.de} \\ 
}

\begin{document}

\maketitle

\begin{abstract}
We study the training dynamics of neural classifiers through the lens of binary hypothesis testing. We re-formalize classification as a collection of binary tests between class-conditional distributions induced by learned representations and show empirically that, along training trajectories, well-generalizing networks progressively approach Neyman–Pearson optimal decision rules, as measured by monotonic growth in the KL divergence retained by learned representations. We provide sufficient conditions for exact optimality, discuss its implications for training regularization, and define an informational plane (so-called Evidence-Error plane) where convergence can be assessed methodically across network architectures.

\end{abstract}

\section{Introduction}
Supervised classification can be viewed fundamentally as a statistical hypothesis testing problem: given input samples $x \sim \mathcal{X}$ and associated labels $y \sim \mathcal{Y}$ such that there is a non-trivial conditional relation between the two, the task of a classifier is to decide, for each observation $x$, between competing hypotheses corresponding to class-conditional distributions $\{\mathbb{P}(X | Y = y) \ | \ \forall y\in\mathcal{Y}\}$. 

Classical decision theory characterizes the optimal solution to this, in the binary case, via likelihood ratio tests and the Neyman–Pearson (NP) lemma, which establishes optimal trade-offs between type-I and type-II errors in terms of likelihood ratios and Kullback–Leibler (KL) divergence \citep{kullback1951information}. Despite the widespread success of deep neural networks as classifiers, the relationship between modern network training and classical hypothesis testing remains poorly understood. In particular, it is unclear whether backpropagation of cross-entropy loss implicitly steers learned representations toward statistically optimal decision regions. 

Our contributions in closing this gap are as follows:
\begin{itemize}[label=--, leftmargin=10pt, itemsep=2pt, topsep=2pt]
    \item We formalize supervised classification as a sequence of binary hypothesis tests on learned representations, and provide \textbf{sufficient conditions} under which cross-entropy training yields representations that are sufficient statistics for the data.

    \item We introduce a controlled synthetic classification dataset with analytically known class-conditional KL divergence, enabling comparison between learned decision rules and statistically optimal tests.

    \item We develop an informational plane (the so-called \textit{Evidence-Error Plane}) in which all networks are mapped to points within an achievable region; allowing direct comparison across architectures as well as tracking the time evolution of a network.

    \item We empirically show that, for different architectures and (both toy and real) datasets, training dynamics of classifier networks trace those of induced likelihood ratio tests between class-conditional feature distributions in the defined plane.
\end{itemize}

\section{Related Work}
The gap between the complexity of modern neural networks and principled tools for analyzing their behavior has been a topic of recent discussion \citep{zhang2021survey}. The Information Bottleneck (IB) was proposed as a global theory of learning dynamics; framing training as a trade-off between predictive information $I(T;Y)$ and representational complexity $I(X;T)$ \citep{shwartz2017opening} with $T$ denoting the network output. However, IB has been criticized on two main grounds: (i) for deterministic networks, $I(X;T)$ is typically ill-posed (constant or infinite); and (ii) mutual information measurements rely on high-dimensional histogram density estimates, which are known to be unreliable \citep{saxe2019information},  \citep{kolchinsky2018caveats}. These issues have motivated extensions that replace or refine the tracked quantities, including generalized-IB and other partial information decompositions \citep{westphal2025generalized}, \citep{tax2017partial}.

In contrast, explainable AI (xAI) methods focus on local interpretability, producing per-sample explanations via sensitivity analysis, local surrogate models, or feature attributions (see e.g. \citep{montavon2019layer}, \citep{ribeiro2016should}). While diagnostically useful, such methods do not define a distribution-level object and lack a ground-truth notion of correctness \citep{rawal2025evaluating}. 

Thus, information-theoretic quantities remain attractive for global analysis: they are invariant under invertible reparameterizations and admit operational interpretations in coding, prediction, and statistical decision-making. Their usefulness, however, depends critically on whether the quantities are well-defined for modern architectures and can be estimated reliably --- precisely the issues that motivate alternative formulations and estimators beyond classical IB \citep{saxe2019information}.

\citep{anders1999model} proposes a model selection procedure for neural networks based on hypothesis testing, (as opposed to, for example, cross validation) by iteratively expanding network topology and performing hypothesis tests on whether the added units correlate with residuals of the original network. Similarly \citep{mandel2024permutation} suggests a hypothesis testing framework for input feature attribution, akin to local xAI methods. Finally, we note \citep{barreno2007optimal}, where the NP lemma is used to find combinations of multiple binary classifiers that achieve optimal error rate trade-offs, as characterized by the ROC curve. However, to the best of our knowledge, directly interpreting a neural network as a hypothesis test is not an approach pursued before.

\section{Theoretical Background}

In this section we outline the theoretical framework of binary hypothesis testing and its connection to neural network representations.

\subsection{Binary Hypothesis Testing}

Let $Z$ be a random variable taking values in a measurable space $(\mathcal{Z}, \mathcal{B})$. Consider two hypotheses
\begin{align}
    H_0 : Z \sim P_0 \qquad \text{vs.} \qquad H_1 : Z \sim P_1,
\end{align}
where $P_0, P_1$ are probability measures on $(\mathcal{Z}, \mathcal{B})$
with $P_1 \ll P_0$. A (possibly non-deterministic) test is a measurable function $\phi: \mathcal{Z} \to [0,1]$, where $\phi(z)$ denotes the probability of rejecting
$H_0$ given observation $Z = z$. The type-I and type-II error rates of $\phi$ are defined respectively as
\begin{align}
    \alpha(\phi) := \mathbb{E}_{P_0}[\phi(Z)], \qquad
    \beta(\phi)  := \mathbb{E}_{P_1}[1 - \phi(Z)].
\end{align}
The Neyman--Pearson Lemma \citep{neyman1933ix} states that for any fixed level $\alpha \in (0,1)$, the log-likelihood ratio test (where $\mathbf{1}$ denotes the indicator function)
\begin{align}
    \phi^*(z) := \mathbf{1}\!\left[\log\frac{dP_1}{dP_0}(z) > \tau\right],
\end{align}
 with $\tau \in \mathbb{R}$ chosen so that $\alpha(\phi^*) = \alpha$, is
\textbf{uniformly most powerful}: for any other test $\phi$ with $\alpha(\phi) \leq
\alpha$, we have $\beta(\phi^*) \leq \beta(\phi)$.

To study the asymptotic behaviour of this test, suppose $n$ i.i.d.\ observations $Z^n =
(Z_1, \ldots, Z_n)$ are drawn from either $P_0^{\otimes n}$ or $P_1^{\otimes
n}$, the $n$-fold product distributions of $P_0, P_1$. For a test $\phi_n$ using $n$ samples with type-I error $\alpha(\phi_n) \leq
\alpha_n$ for some fixed $\alpha_n \in (0,1)$, define the optimal type-II
error exponent
\begin{align}
    \beta_n^*(\alpha_n) :=
    \inf_{\phi_n:\,\alpha(\phi_n)\leq\alpha_n}
    \beta(\phi_n).
\end{align}
Stein's Lemma \citep{chernoff1956large} establishes that the KL
divergence governs the optimal error exponent:
\begin{align}\label{Stein}
    \lim_{\epsilon \to 0}\,\lim_{n \to \infty}
    -\frac{1}{n}\log\beta_n^*(\epsilon)
    = D_{\mathrm{KL}}(P_1 \| P_0)
    = \mathbb{E}_{P_1}\!\left[\log\frac{dP_1}{dP_0}\right].
\end{align}

\subsection{Neural Networks as Likelihood Ratio Estimators}

Let $(\mathcal{X},\mathcal{F}_{\mathcal{X}})$ be a measurable space with
measure $\mu : \mathcal{X} \to \mathbb{R}^+$, and $\mathcal{Y} = \{1,
\ldots, K\}$ be the set of classes. Let $(X, Y)$ be a jointly distributed
random pair with uniform class priors $\mathbb{P}(Y=c) = 1/K, \forall c \in \mathcal Y$. We define class-conditional distributions as
\begin{align}
X_c := \mathbb{P}(X \mid Y=c), \qquad X_{\bar{c}} := \frac{1}{K-1}\sum_{c' \neq c}
X_{c'}
\end{align}
$X_{\bar c}$ stands for the complement distribution of class $c$ and can be written alternatively as $X_{\bar c} = \mathbb P(X | Y \neq c)$. Let $\Theta$ be a class of measurable functions (i.e.\ neural networks) $\theta : \mathcal{X} \to \mathbb{R}^K$, with softmax output
\begin{align}
    \sigma_c(\theta(x)) := \frac{\exp(\theta(x)_c)}{\sum_{c'}\exp(\theta(x)_{c'})},
\end{align}
and let $\eta_c(x) := \mathbb{P}(Y=c \mid X=x)$ denote the true class
posterior. Define joint distributions
\begin{align}
    P(X=x,\, Y=c) = \tfrac{1}{K}\, X_c(x), \quad
    Q(X=x,\, Y=c) = \tfrac{1}{K}\, X_{\bar{c}}(x)
\end{align}
inducing the hypothesis test 
\begin{align}
H_0 : (X,Y) \sim P \ \text{vs.}\ H_1 : (X,Y) \sim Q
\end{align}

Intuitively, this test asks given a sample $x$ and class $c$, whether $x$ belongs to $c$ or to any other class. By Neyman--Pearson, the optimal test statistic is the LLR
\begin{align}
\Lambda_c(x) = \log\frac{X_c(x)}{X_{\bar{c}}(x)}
\end{align}

which in high-dimensional settings is intractable to estimate from samples directly \citep{wang2019nonparametric}. Therefore a direct application of the NP test is impractical. 

Our central thesis is that \textit{a neural network $\theta$ learns a discriminative surrogate for $\Lambda_c(x)$ on a lower dimensional representation space (where the output logits reside) and bypasses direct density estimation.}

We write the \textit{input divergence} $D_{\mathrm{inp}}$ as the KL divergence between $P$ and $Q$:
\begin{align}
    D_{\mathrm{KL}}(P \| Q)
    = \frac{1}{K}\sum_{c=1}^K D_{\mathrm{KL}}(X_c \| X_{\bar{c}})
    =: D_{\mathrm{inp}},
\end{align}
which defines a universal performance ceiling for the task. Since the classification layer operates on the representation $Z = \theta(X)$, with class-conditional distributions defined similarly $Z_c := \mathbb{P}(Z \mid Y=c)$ and
$Z_{\bar{c}} := \mathbb{P}(Z \mid Y \neq c)$, the data processing inequality
(DPI) \citep{polyanskiy2016information} gives
\begin{equation}\label{DPI}
    D_{\mathrm{KL}}(Z_c \| Z_{\bar{c}}) \leq D_{\mathrm{KL}}(X_c \| X_{\bar{c}}),
    \quad \forall c \in \mathcal Y,
\end{equation}
so that we have the \emph{classifier divergence}
\begin{equation}\label{Network Div}
    D_\theta := \frac{1}{K}\sum_{c=1}^K D_{\mathrm{KL}}(Z_c \| Z_{\bar{c}})
    \leq D_{\mathrm{inp}}.
\end{equation}

The \emph{divergence gap} $\delta(\theta) := D_{\mathrm{inp}} - D_\theta \geq 0$ intuitively measures how much discriminative information is lost in the representation.

Alongside it, we define the \emph{cross-entropy gap}
\begin{equation}
    \epsilon(\theta) := \mathcal{L}(\theta) - H(Y|X) \geq 0,
\end{equation}
where $\mathcal{L}(\theta) := -\mathbb{E}_{(X,Y)}[\log\sigma_Y(\theta(X))]$
is the cross-entropy loss and $H(Y|X) := -\mathbb{E}_{(X,Y)}[\log\eta_Y(X)]$
is the irreducible conditional entropy of $Y|X$. One can show (see
Appendix~\ref{app:epsilon_kl}) that
\begin{equation}\label{eq:epsilon_kl}
    \epsilon(\theta) = \mathbb{E}_X\!\left[
        D_{\mathrm{KL}}\!\left(\eta(X) \,\|\, \sigma(\theta(X))\right)
    \right],
\end{equation}
which formalizes the interpretation of $\epsilon(\theta)$ as the expected pointwise KL divergence between the true posterior and the network’s softmax outputs; a measure of their deviation.

A natural question is when $\delta(\theta) = 0$, i.e.\ when the representation fully preserves the input divergence and the network
achieves the Stein limit~\eqref{Stein}. The following theorem gives a sufficient condition in terms of cross-entropy minimization.

\begin{theorem}[Exact KL Preservation]\label{thm:exact_kl}
    Let $\theta^* \in \arg\min_{\theta \in \Theta}\,\mathcal{L}(\theta)$ be a network with minimal cross-entropy. Under Assumptions~\ref{ass:realizability}
    (Realizability) and~\ref{ass:bounded_posteriors} (Bounded Posteriors), we have both 
    \begin{align}
        \epsilon(\theta^*) = 0 \ \text{and} \ \delta(\theta^*) = 0.
    \end{align}
    Moreover, for each $c \in \mathcal{Y}$, the representation $\theta^*(X)$
    is a sufficient statistic for the pair $(X_c, X_{\bar{c}}): \exists \, h_c : \mathbb{R}^K \to \mathbb{R}$ measurable such that
    \begin{equation}
        \Lambda_c(x) = h_c(\theta^*(x)) \qquad \mu\text{-a.e.}
    \end{equation}
\end{theorem}

The assumptions and full proof are given in Appendix~\ref{app:kl_preservation}.
Intuitively, realizability ensures the function class is rich enough to
recover the true posterior exactly, at which point the network's logits
encode the LLR $\Lambda_c$ and no discriminative information is lost.
The theorem thus establishes that cross-entropy minimization is a sufficient
mechanism for a network to operate at the Stein limit, provided the architecture is general enough for the task at hand. 

\subsection{Evidence--Error Plane}

We define a 2-D coordinate system (analogous to the Information Plane from \citep{shwartz2017opening}) where networks can be translated into points in terms of their type-II error rates versus retained KL divergence. Formally, each network gets mapped to a point
\begin{equation}\label{EE plane}
    \Phi(\theta) = (P_\theta,\, D_\theta) \in \mathbb{R}^2_+,
\end{equation}
where $P_\theta = -\log\beta_\theta$ is the minus-log empirical type-II error
rate and $D_\theta$ is as in~\eqref{Network Div}. This disentangles
\emph{representation quality} (divergence retained; the $y$-axis) from
\emph{classification efficiency} (error achieved; the $x$-axis).

The diagonal $P_\theta = D_\theta$ corresponds to the asymptotic Stein limit; networks above it would violate~\eqref{Stein} and therefore cannot exist. Networks well below the diagonal are considered \textit{information-inefficient}. More concretely, we characterize this inefficiency with the following two cases:

\begin{enumerate}[label=\alph*), leftmargin=10pt, itemsep=2pt, topsep=2pt]

    \item For a network \( \theta \); if \( P_\theta < D_\theta \) for high \( D_\theta \), \( \theta \) is information inefficient, because it is not able to convert all the extracted divergence into performance.

    \item Given two networks \( \theta_1, \theta_2 \); if \( D_{\theta_1}  > D_{\theta_2} \) but \( P_{\theta_1} \approx P_{\theta_2} \), \( \theta_2 \) is information inefficient, because although it achieves the same performance as \( \theta_1 \), it is not able to extract all the \textit{usable} divergence from the data that \( \theta_1 \) could. Note that for the same reason above, \( \theta_1 \) is also information inefficient (as it has \( P_{\theta_1} < D_{\theta_1} \) necessarily).

\end{enumerate}

On the other hand, the DPI bound $D_\theta \leq D_{\mathrm{inp}}$ caps the $y$-axis, so
the achievable region $\mathcal A$ for \textit{all networks} is formed by the triangle
\begin{equation}\label{EE region}
    \mathcal{A} = \{(P_\theta, D_\theta) \in \mathbb{R}^2_+ :
    0 \leq P_\theta \leq D_\theta \leq D_{\mathrm{inp}}\}.
\end{equation}

The set $\mathcal A$ also defines a "fully convergent"  --- one satisfying Theorem~\ref{thm:exact_kl} --- network, namely on the point $(D_{inp}, D_{inp})$; the top right corner of the triangle. Figure ~\ref{fig:evidence-error-binary-image} is an example of a network existing on the optimal Stein line and approaching to the full convergence point of $(D_{inp}, D_{inp})$.

In practice, network trajectories through $\mathcal{A}$ during training provide a finer-grained convergence diagnostic than accuracy curves alone. Since KL estimation over
the logit space has negligible computational cost compared to a training epoch, (e.g the kNN based estimator we utilize has $O(k\log n)$ cost with a $k$-d tree implementation, see \ref{app:kNN-KL}) the evidence--error plane can be utilized as a lightweight wrapper around any training loop to asses convergence.

\section{Experimental Results}
\begin{figure}
    \centering
    \includegraphics[width=0.7\linewidth]{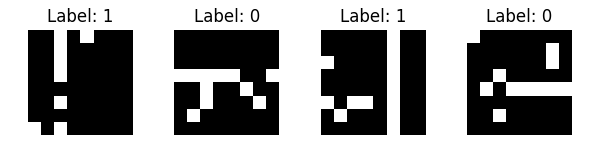}
    \caption{\textbf{Sample points from the Binary Image Dataset}}
    \label{fig:bin-img-samples}
\end{figure}
We train different networks with comparable sizes/architectures and study their trajectories on the Evidence-Error plane. Each network consists of 4 hidden layers (64-32-16-8 neurons) and a softmax classifier. All layers are fully connected and hidden layers have ReLU nonlinearities. The training is done via cross entropy loss on the logits, optimized via Adam (with learning rate $\gamma=0.001$ and $\beta_1 = 0.9, \beta_2 = 0.999$) \citep{kingma2017adam}. Each network is trained for 50 epochs. All experiments are implemented with the PyTorch framework \citep{paszke2019pytorch}.

We estimate class-conditional KL divergences, both for the input distributions $D_{\mathrm{KL}}(X_c \| X_{\bar{c}})$ and representations $D_{\mathrm{KL}}(Z_c \| Z_{\bar{c}})$ using a k-nearest-neighbor (kNN)–based estimator, which approximates local density ratios from sample neighborhoods without any density modeling \citep{perez2008kullback}. Although it exhibits finite sample bias, the kNN estimator is asymptotically convergent (for any $k$) and in fact, known to be nearly minimax sample rate optimal in mean square error \citep{zhao2020minimax}. Coupled with the low computational demand, we deem it to be the ideal estimator for our application. All KL estimates are computed on full representation vectors (logit outputs of the classifier layer, before softmax is applied) using identical settings across layers and epochs. Appendix \ref{kNN KL Bias} describes a heuristic for choosing $k$, among other bias reduction methods we employ.

\subsection{Multivariate Gaussians}\label{Gaussian Experiment}
Before moving on to real (or realistic) datasets, we first study the case when the two classes are Gaussians as a sanity check. We generate two overlapping multivariate Gaussians with equal priors, unit covariance and a mean shifted only by one in the first coordinate. Formally: $X_0 \sim \mathcal{N}(\mu_0, I_d)$ and $X_1 \sim \mathcal{N}(\mu_1, I_d)$ with $\mu_0 = (0,0, ..0) \text{ and } \mu_1 = (1,0, ..0)$. Analytic KL divergence is symmetric and known in this case; $D_{\mathrm{KL}}(X_0 \| X_1) = D_{\mathrm{KL}}(X_1 \| X_0) = 0.5$ nats $\approx 0.721$ bits.

Moreover, since they have unit covariance, the LLR test and associated error probabilities are:
\begin{align}
&(\mu_1 - \mu_0)^{\top} X \geq \gamma \\
\beta^*(\alpha) &= \Phi(\Phi^{-1}(1 - \alpha) - \norm{\mu_1 - \mu_0}_2)
\end{align} 
where $\norm{\mu_1 - \mu_0}_2 = 1$ and $\Phi(x)$ is the CDF of the standard normal distribution. 

\begin{figure}
    \centering
    \includegraphics[width=0.5\linewidth]{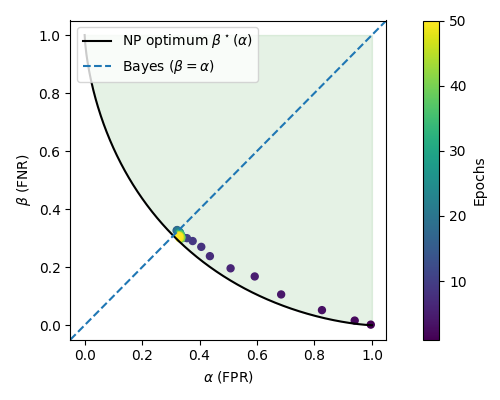}
    \caption{\textbf{4D Gaussian Dataset Error Rates:} Green region is all achievable error rates and the black curve defines the NP-optimal envelope. Each point is the empirical error rates of the network on the corresponding epoch. The network converges to the Bayes optimal classifier's error rate, which is the intersection between the curve and the dotted line $\beta = \alpha$. See \ref{Gaussian Experiment}.}
    \label{fig:gaussian-np}
\end{figure}

Figure \ref{fig:gaussian-np} plots the empirical $(\alpha, \beta)$ error rates of a network on this dataset versus the NP optimal error envelope, $\beta^*(\alpha)$. The network exists nearly on the boundary and approaches the point intersecting the NP-optimal curve and the line corresponding to the Bayes error, $\beta = \alpha$ (since they have equal priors). This implies that the trained network is equivalent to the Bayes optimal classifier. Section \ref{Conclusion} discusses this fact more in detail.

\subsection{Networks on the Error-Evidence Plane}

We move on to track the evolution of the same network architecture on the following datasets:
\begin{enumerate}
    \item \textbf{Binary Image:} A custom toy dataset with binary classes and $d\times d$ images consisting of binary pixels, corrupted through a Binary Symmetric Channel \citep{polyanskiy2016information} with bit flip probability $p$. See Appendix \ref{Binary Image Dataset Appendix} for a detailed description of the data generation and KL divergence calculation. Sample images for $d=8$ can be found in Figure \ref{fig:bin-img-samples}.
    \item \textbf{Yin-Yang:} A three class toy dataset (yin, yang and dot) generated by $(x,y)$ coordinates on the unit disk. The coordinates $(1-x, 1-y)$ are also used as input features for symmetrizing the input \citep{kriener2022yin}.
    \item \textbf{MNIST:} The MNIST handwritten digits dataset, consisting of $28\times 28$ grayscale images from $10$ classes \citep{lecun1998mnist}.
    \item \textbf{CIFAR-10:} A natural image dataset, consisting of $32\times 32$ RGB images with $10$ classes \citep{krizhevsky2009cifar}.
\end{enumerate}

\begin{figure}
    \centering
    \begin{subfigure}[t]{0.48\linewidth}
        \centering
        \includegraphics[width=\linewidth]{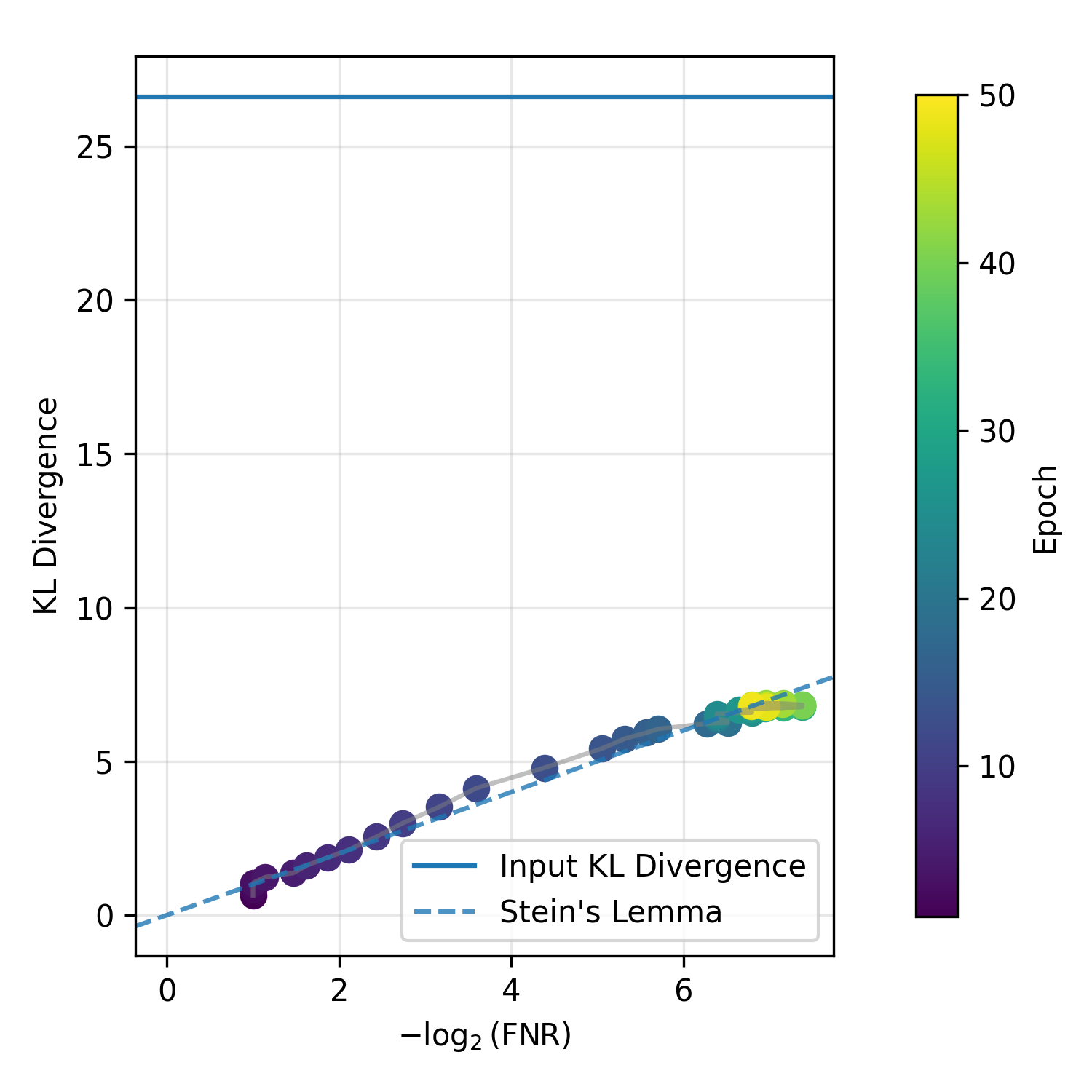}
        \caption{\textbf{Binary Image:} KL divergence between class-conditioned representations improves with training, driving the network exactly to the Stein error regime -- the prototypical NP-optimal behavior. Solid blue line is $D_{inp}$; dotted is Stein's Lemma prediction.}
        \label{fig:evidence-error-binary-image}
    \end{subfigure}
    \hfill
    \begin{subfigure}[t]{0.48\linewidth}
        \centering
        \includegraphics[width=\linewidth]{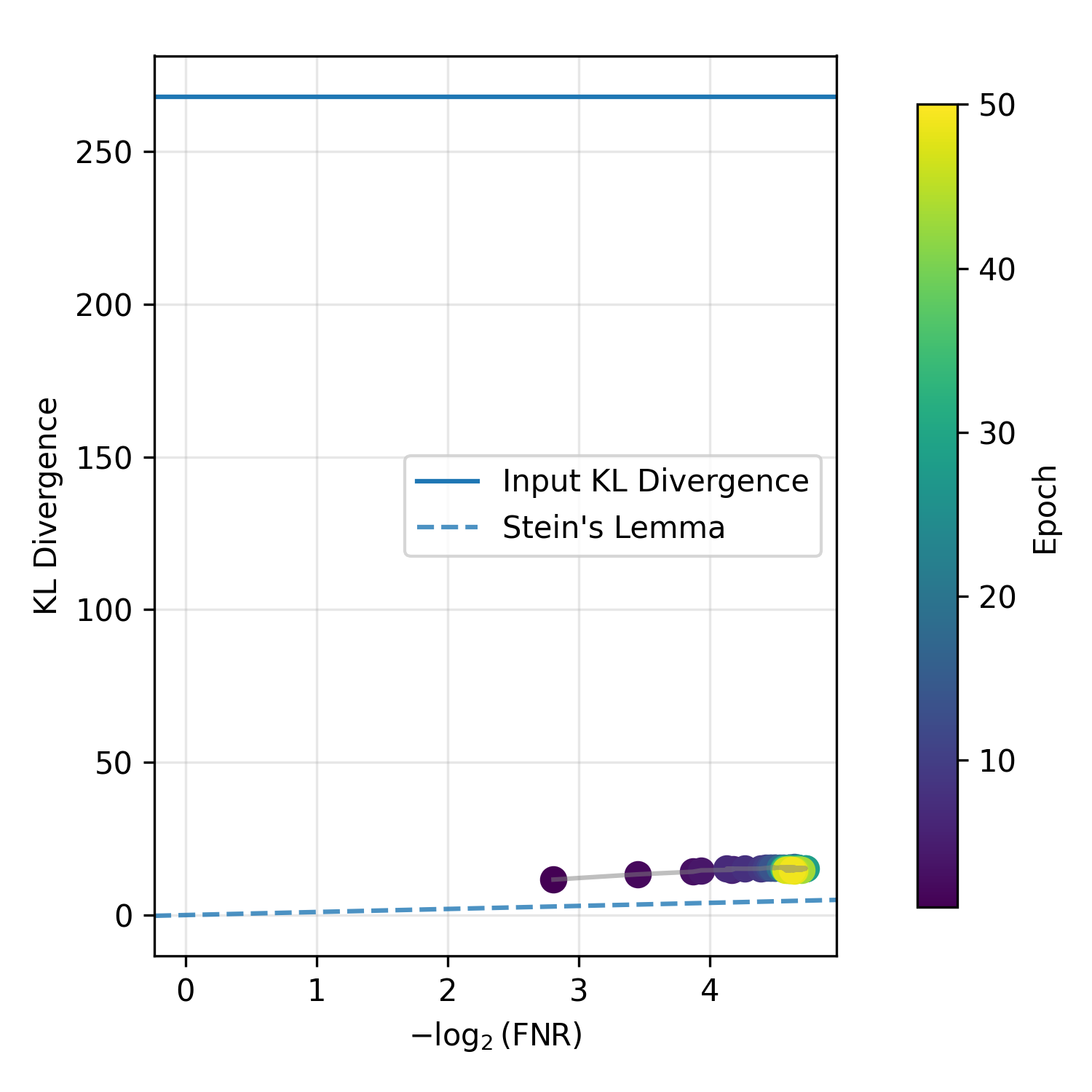}
        \caption{\textbf{MNIST:} Same network trained on MNIST. A similar trajectory is observed, although the network operates noticeably worse than the Stein limit.}
        \label{fig:evidence-error-mnist}
    \end{subfigure}
    \caption{\textbf{Comparison of DNN behavior on Binary Image and MNIST datasets.}}
    \label{fig:evidence-error-combined}
\end{figure}

Figure \ref{fig:evidence-error-binary-image} presents the network trained on the Binary Image dataset with $d=8$ and $p=0.1$. This results in a relatively easy task that the network is able to solve in a handful of epochs. We observe that the network also converges, in the sense suggested by Theorem~\ref{thm:exact_kl}. Figure \ref{fig:evidence-error-mnist} shows the same network trained on MNIST, with a similar trajectory; although it is information inefficient, as it is on average $\sim 10.4$ bits above the Stein limit. 

\subsection{$n > 1$ case: Majority Voting Classification}
As argued before, Stein's lemma only characterizes the $n \to \infty $ regime, where many i.i.d samples are available at inference time. However, typical neural networks operate at $n=1$; because only a single input sample is processed for classification. To understand the effect of $n > 1$ samples being available at inference, we conduct the following experiment:

We train a network as usual, then generate $x = \{x_1, x_2, ..., x_n\}$ i.i.d input samples all belonging to the same class $c$. At inference, we collect the network output on all samples $\Theta(x) = \{\theta(x_1), \theta(x_2) ... \theta(x_n)\}$ and perform a majority voting to determine class output:
\begin{equation}
    \hat{y} = argmax_{k\in\{1,\dots,K\}} \sum_{i=1}^n \mathbf{1}\{ \theta(x_i)=k\}.
\end{equation}

Then all $x_i$ are classified as $\hat{y}$ and error rates $(\alpha, \beta)$ are calculated as the result of that classification. 

\begin{figure}
    \centering
    \includegraphics[width=0.6\linewidth]{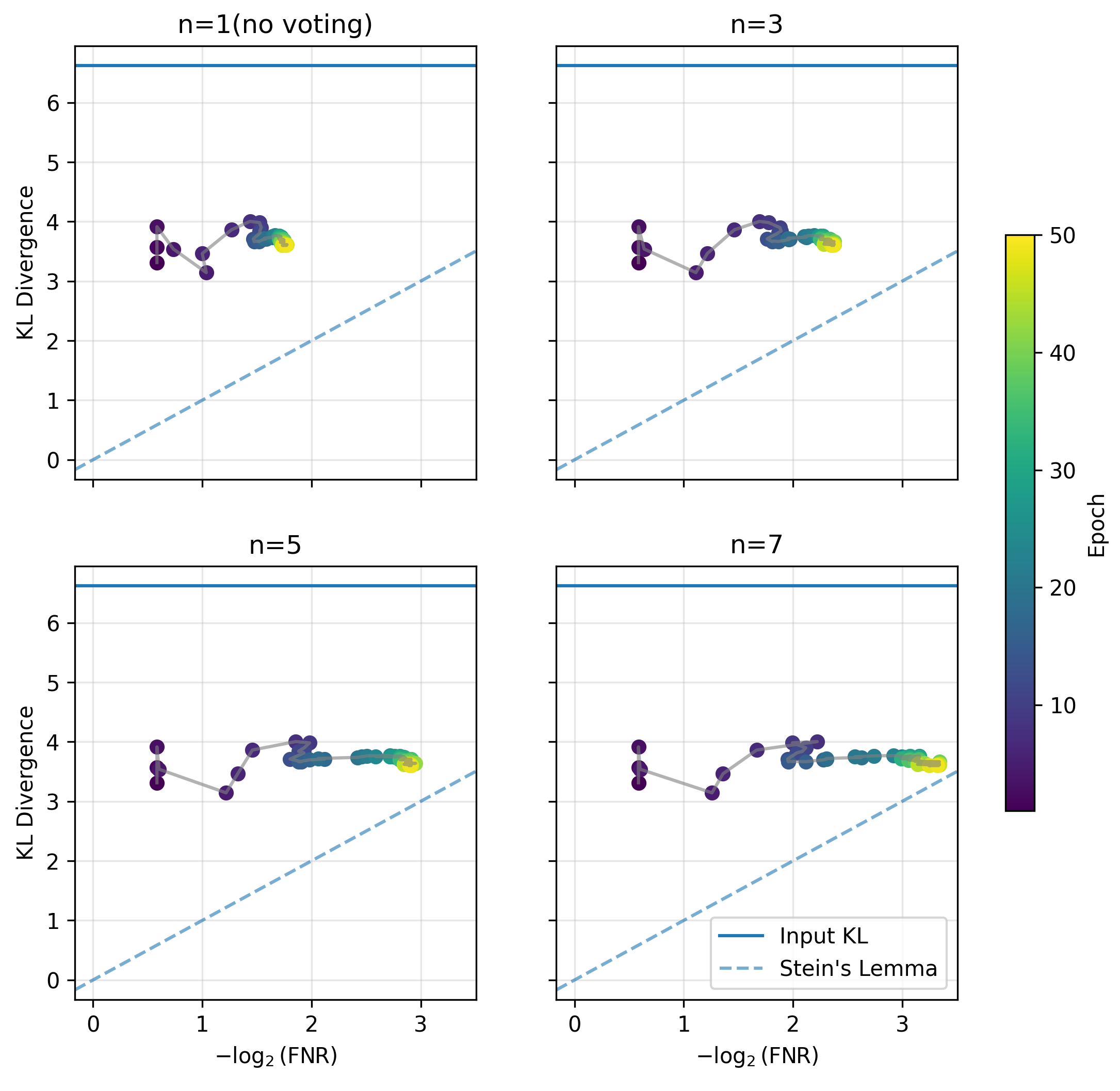}
    \caption{\textbf{Majority Voting on Information Inefficient Network:} A smaller network with only two hidden layers, trained on the Yin-Yang dataset and evaluated with majority voting classification. As the number of samples $n$ used in the vote increase, the network performance also increases while $D_\theta$ stays constant.}
    \label{fig:yin-yang-voted}
\end{figure}

A priori, it is not the case that this would be equivalent to the LLR test on $n$ samples. Moreover, it is not even true that error rates necessarily decline monotonically with $n$ (see Appendix \ref{Maj vote counterex} for a counterexample). However, we observe that for networks that we have classified as "information inefficient" ($D_{\theta} \gg P_{\theta})$, majority voting improves $P_\theta$ and allows the network to approach the Stein regime (See Figure \ref{fig:yin-yang-voted}). Indeed, in early epochs when sufficient $D_{\theta}$ is not reached, majority voting does not provide a significant increase in $P_{\theta}$; whereas in later epochs the network moves significantly right on the $P_{\theta}$ axis. 

\subsection{Spiking Neural Networks} \label{SNN Section}
A separate class of neural networks we study are Spiking Neural Networks (SNNs), which implement discrete-time Leaky Integrate-and-Fire (LIF) neurons rather than standard perceptrons and are trained via surrogate gradient methods \citep{eshraghian2023training}. Each input sample $x_i$ is encoded, via a fixed encoding scheme, into a binary spike train with $S_0[t] \in \{0,1\}^{d}$ of length $\tau \in \mathbb{N}, t\in\{1,2,...\tau\}$. Neurons maintain an internal state referred to as the \textit{membrane potential} $U_l[t]$, where $l$ indexes the network layer. The membrane potential evolves according to
\begin{align}
    U_l[t] &= \eta U_l[t-1] + W_{l-1} S_{l-1}[t] - V_{\mathrm{th}}S_l[t-1], \\
    S_l[t] &= \Theta\!\left(U_l[t] - V_{\mathrm{th}}\right)
\end{align}
where $\eta \in (0,1)$ is the leak factor, $W_{l-1}$ denotes the synaptic weight matrix from layer $l-1$ to layer $l$, $V_{\mathrm{th}}$ is a fixed firing threshold, and $\Theta(\cdot)$ is the Heaviside step function. After a neuron fires, its potential is reset by $V_{\mathrm{th}}$. 

This event-driven dynamics lead to sparse activations and when deployed on neuromorphic hardware, enables substantial reductions in inference-time energy consumption \citep{kudithipudi2025neuromorphic}. Due to the explicit notion of internal state given by the membrane potential, we analyze SNNs with the same architecture as their non-spiking counterparts and define class-conditional representations via $U_c = \mathbb{P}(U \mid Y = c)$ with $U_{\bar{c}}$ defined analogously. 

We encode inputs via rate-coding using Poisson spike trains, where for each feature $x_i$, the spikes are sampled i.i.d. with $S_i[t]\sim\mathrm{Bernoulli}(x_i)$ for $\tau=5$ assuming normalized inputs. We take $\eta = 0.95$ constant and we train our model with the SNNTorch framework, using  $S \approx \frac{1}{\pi}arctan(\pi U)$ as the surrogate for backpropagation. \citep{eshraghian2023training}.

\begin{figure}[t]
    \centering
    \includegraphics[width=0.5\linewidth]{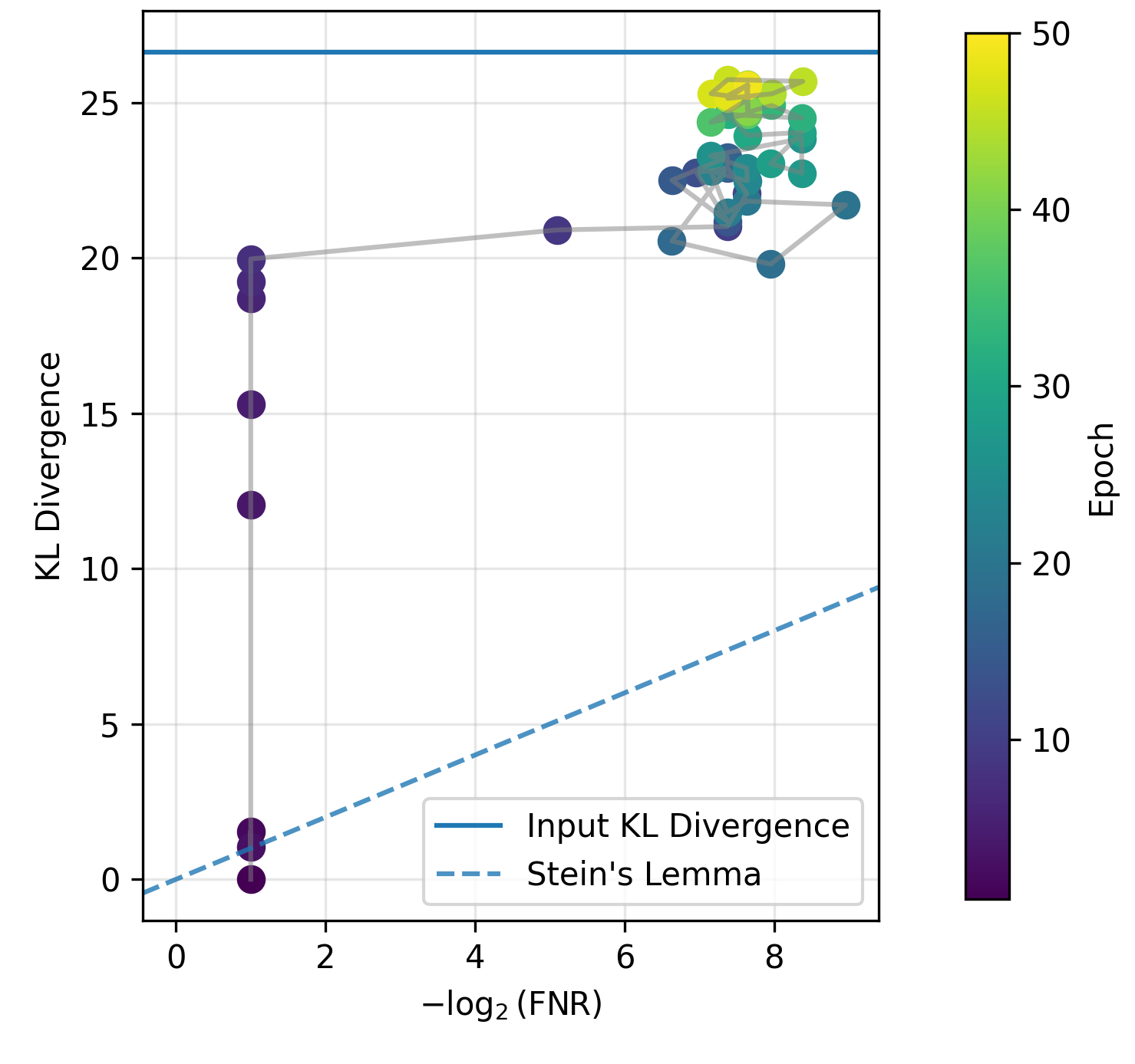}
    \caption{\textbf{SNN trained on Binary Image:} A dramatic increase in classifier divergence in early epochs followed by a "fitting" phase of performance increase in later epochs. See \ref{SNN Section}.}
    \label{fig:evidence-error-snn}
\end{figure}

Figure \ref{fig:evidence-error-snn} shows the Evidence-Error plane trajectory of the SNN on Binary Image. In contrast to DNN's, we observe a two-phase trajectory (analogously to Information Bottleneck \citep{shwartz2017opening}, although "reversed"); where the early epochs have constant error rates but dramatically increase classifier divergence (well above the equivalent DNN), and later epochs exploit that divergence by increasing $P_{\theta}$. 

However, for fixed weights $W_{l}$, note that the support of the membrane potential distribution is effectively a finite alphabet, as updates to the potential are done in discrete steps. Therefore the estimator may be unreliable (see Appendix \ref{SNN Finite Alphabet} for a detailed argument). 

\subsection{ResNet50 on CIFAR-100: Scaling to Realistic Scenarios}

We study a larger classifier on a more challenging dataset; namely the ResNet-50 architecture \citep{he2016deep} on CIFAR-100 \citep{krizhevsky2009cifar}. Unlike other datasets, the kNN KL divergence estimator  requires subsampling here, due to class imbalance: the out-class distribution $X_{\bar c}$ contributes 99 times as many samples as the in-class distribution $X_c$, Without correction, this imbalance biases the estimator toward negative values, as kNN distances collapse. Additionally, for $D_{\mathrm{inp}}$, we apply a preliminary PCA (with 256 components) to reduce the dimensionality, as each class provides only 100 samples against the 3072-dimensional 
input space. By the DPI, this can only reduce the true $D_{\mathrm{inp}}$, but it allows for a meaningful lower bound to be drawn.

\begin{figure}[h]
    \centering
    \begin{minipage}[c]{0.48\textwidth}
        \centering
        \includegraphics[width=\linewidth]{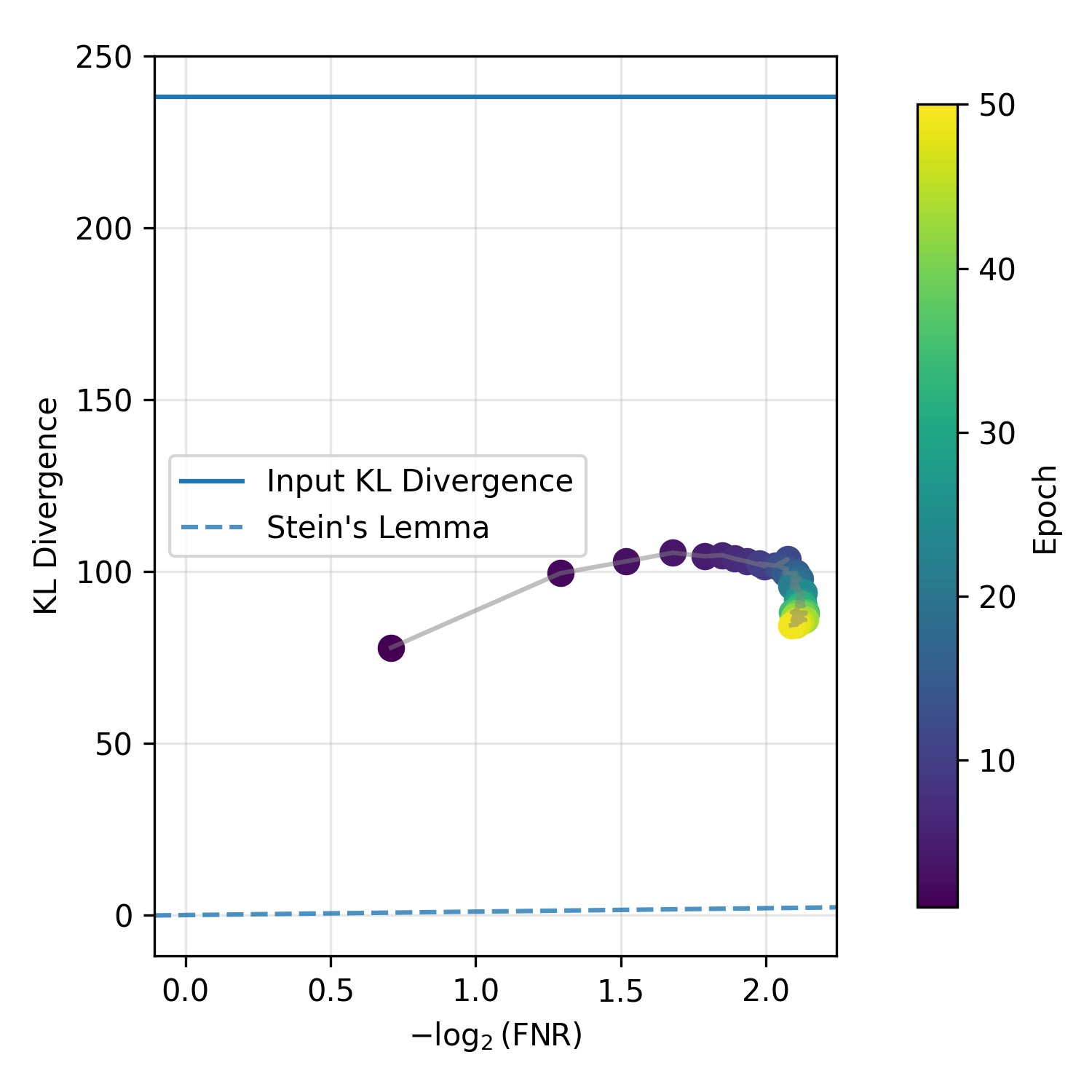}
    \end{minipage}
    \hfill
    \begin{minipage}[c]{0.48\textwidth}
        \centering
        \includegraphics[width=\linewidth]{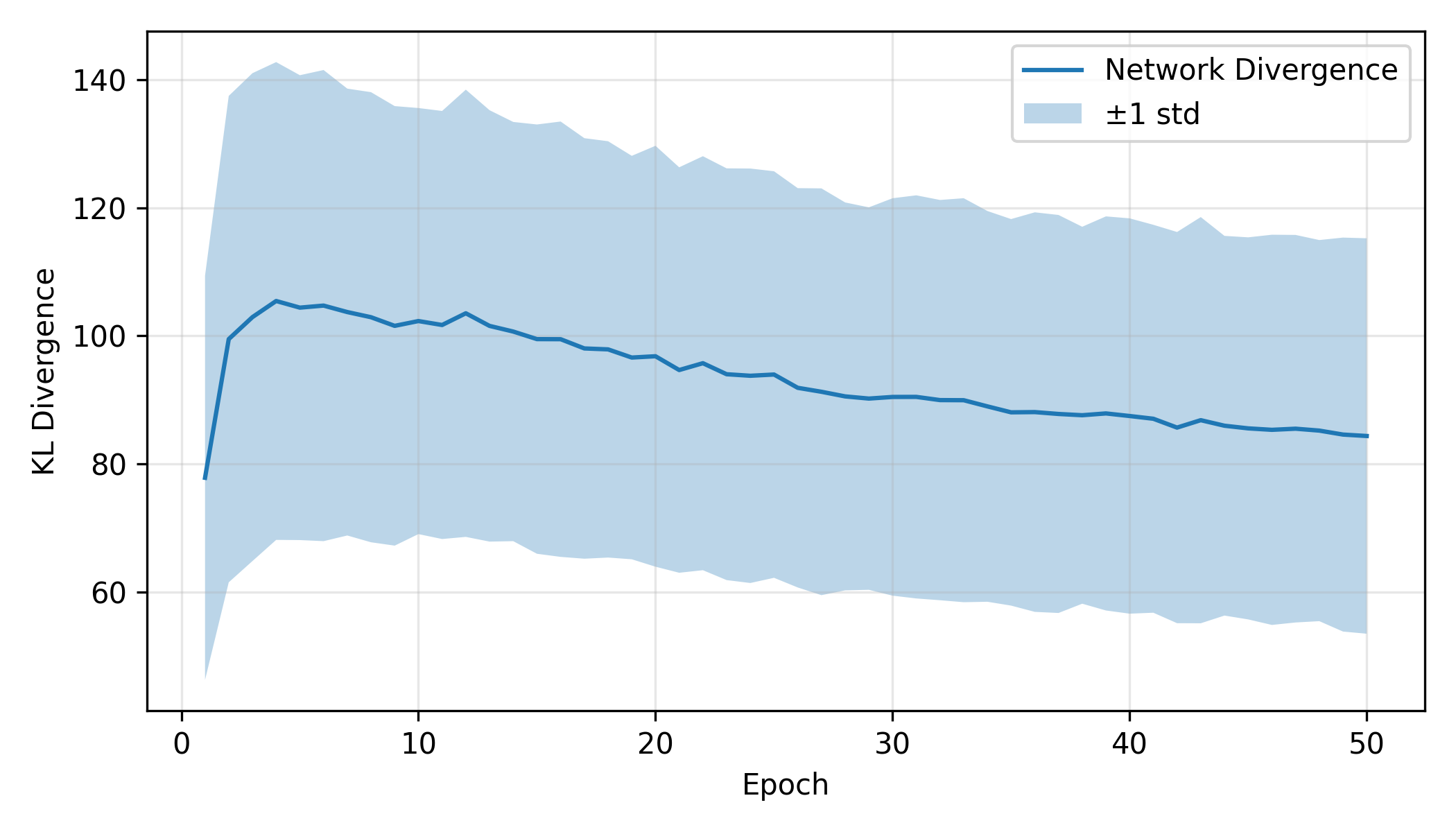}
    \end{minipage}
    \caption{\textbf{ResNet50 on CIFAR100}: We observe, unlike other settings, a drop in KL divergence later in training. The plot on the right shows the evolution of $D_\theta$ only and the shaded area shows the standard deviation of $D_{\mathrm{KL}}(Z_c\|Z_{\bar c})$ across classes $c$.}
    \label{fig:cifar100}
\end{figure}

Interestingly in this setting, we observe a declining trend in $D_{\theta}$, starting around the 10th epoch while the error rates continue to drop (as $P_{\theta}$ increases). This is reminiscent of the two-phase IB dynamics of prediction followed by compression \citep{shwartz2017opening}. Although our theoretical analysis does not conjecture the existence of such phases, this may be expected as the IB terms $I(X; \theta(X))$ and $I(\theta(X); Y)$ are  related to our measures. Indeed our divergence gap $\delta(\theta)$ is measuring discriminative information lost in the representation while the compression term of IB measures the \textit{non-discriminative} information the representation \textit{can} discard.

\begin{figure}[t]
    \centering
    \includegraphics[width=0.5\linewidth]{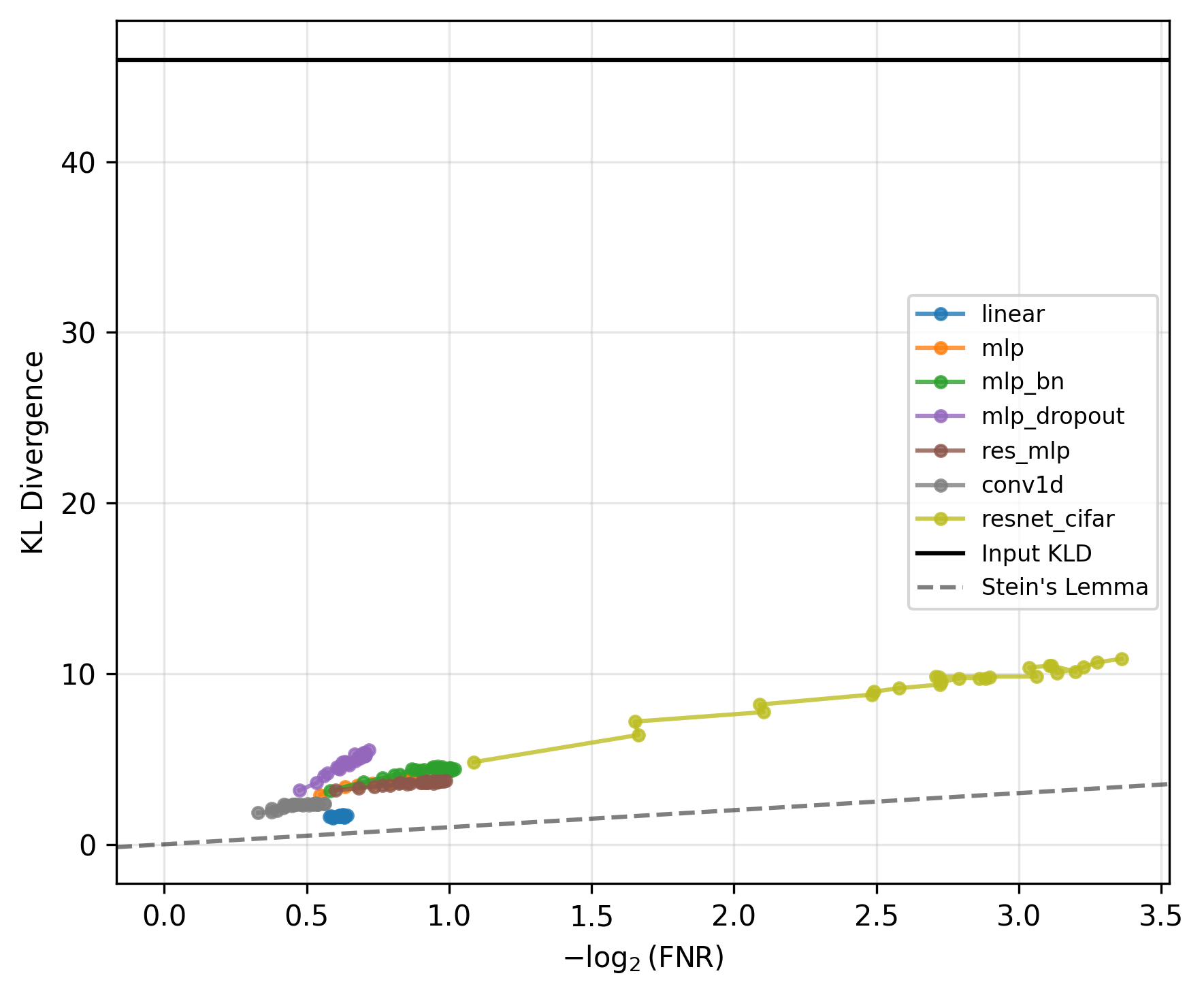}
    \caption{\textbf{Many architectures trained on CIFAR-10}: A variety of architectures trained for 25 epochs on CIFAR-10, placed on the same evidence-error plot. All networks have 4 hidden layers with $64-32-16-8$ neurons respectively and achieve around $50\%$ accuracy, except \textit{resnet\_cifar} which is a ResNet18 \citep{he2016deep} achieving $90.3\%$}
    \label{fig:evidence-error-cifar-all}
\end{figure}

To conclude the experimental section, in Table \ref{tab:architecture-study}, we report $(D_{\theta}, P_{\theta})$ pairs for different trained network architectures, as well as $D_{inp}$ values for the respective datasets. Finally, Figure \ref{fig:evidence-error-cifar-all} shows a variety of architectures trained on CIFAR-10, plotted on the same evidence-error plane.

\begin{table}[t]
  \caption{$(D_{\theta}, P_{\theta})$ values for trained classifiers with different architectures. The most information-efficient network is highlighted. The input divergence $D_{\text{inp}}$ for each dataset is also given as reference.}
  \vspace{0.2cm}
  \label{tab:architecture-study}
  \centering
  \renewcommand{\arraystretch}{1.1}
  \footnotesize
  \setlength{\tabcolsep}{3pt}
  \begin{tabular}{lcccc}
    \toprule
    \textbf{Model Type} & \textbf{Binary Image (26.6)} & \textbf{Yin-Yang (6.3)} & \textbf{MNIST (258.8)} & \textbf{CIFAR10 (45.9)} \\
    \midrule
    Fully Connected (FC) & (7.5, 7.3)  & (4.0, 2.6)  & (14.8, 4.6) & (3.4, 1.0) \\
    FC + BatchNorm       & (7.0, 6.0)  & \textbf{(5.8, 4.2)} & \textbf{(20.7, 5.0)} & \textbf{(4.1, 1.0)} \\
    FC + Dropout         & (7.9, 7.2)  & (4.9, 2.2)  & (29.2, 4.2) & (6.1, 0.8) \\
    FC + Residual Blocks & \textbf{(7.8, 7.4)} & (4.6, 2.4) & (15.9, 4.7) & (3.7, 1.1) \\
    Conv1D               & (6.5, 6.8)  & (3.9, 2.75) & (6.3, 1.4) & (2.2, 0.7) \\
    SNN                  & (25.1, 7.3) &
    (5.0, 1.2)  & (238.0, 4.4) & (7.3, 0.45) \\
    Linear               & (0.9, 1.0)  & (3.7, 1.0)  & (8.9, 3.7) & (1.8, 0.8) \\
    \bottomrule
  \end{tabular}
\end{table}

\section{Conclusion and Future Work} \label{Conclusion}
In this work, we argued theoretically that neural classifier performance can be characterized in terms of binary hypothesis testing and achievable type-II error ($\beta$) based on representation quality. By embedding model trajectories during training into an evidence–error coordinate plane, we disentangled the information available in learned representations from how efficiently it is exploited by the decision rule, and showed that effective training moves networks inside an information-theoretic achievable region, defined by the available divergence between class conditionals and corresponding reachable error rates.

We note however, the main shortcoming of our analysis: in reality, neural network loss functions typically minimize type-I and type-II errors simultaneously. This (Bayesian) setup can be expressed as: 
\begin{equation}
   P_e^* = \min_{\varphi \in \mathcal{T}} \left\{ \pi_P \, \alpha_n(\varphi) + \pi_Q \, \beta_n(\varphi) \right\},
\end{equation}
where $\pi_P, \pi_Q$ are priors of the hypotheses $P$ and $Q$ and $\mathcal T$ is a set of admissible tests. We have briefly observed this in the Gaussian case (see again Figure \ref{fig:gaussian-np}) where the network converges exactly to $P_e^*$. 

For the asymptotic case of $P_e^*$, analogous to Stein's lemma, the Chernoff Information between $P$ and $Q$ defines the optimal error exponent (assuming a common measure $\mu$ for $P$ and $Q$) \citep{chernoff1952measure}:
\begin{align}
\lim_{n \to \infty} \left(-\frac{1}{n} \log P_e^*\right)
=& \sup_{0 \leq s \leq 1}
\left[-\log \int_{\mathcal X} p(x)^s q(x)^{1-s}\, d\mu(x)\right] \\
:=& C(P,Q)
\end{align}
and can equivalently be stated as the following KL divergence:
\begin{equation}
C(P, Q) = D_{\mathrm{KL}}(P_{s^* } \| P) = D_{\mathrm{KL}}(P_{s^* } \| Q)
\end{equation}
where
\begin{equation}
P_{s} (z) = \frac{P^{s}(z)Q^{1 - s}(z)}{\int_{\mathcal X} p(x)^s q(x)^{1-s} d\mu(x)}
\end{equation}
and $s^*$ is the value of $s$ maximizing the first expression. Note that the priors vanish in the asymptotics and $C(P,Q) = C(Q, P)$, unlike KL divergence.

Analytic forms of $C(P,Q)$ beyond a handful of distributions, e.g \citep{nielsen2011chernoff} are not known and estimating it is difficult due to the effective optimization over distributions through $s$. To the best of our knowledge, no reliable kNN estimator even exists. 

Therefore, the Neyman--Pearson analysis presented here should be interpreted as a deliberate simplification, in which the network’s false positive rate is effectively discarded. A natural extension would be to examine whether networks operate near the Chernoff optimal regime in the evidence–error plane and whether the same sufficient conditions hold for cross-entropy training to achieve Chernoff optimality.

\section*{Impact Statement}
We present a deeper theoretical understanding of a broad class of deep neural network models. The analysis provides insight into neural network training dynamics and connects them to the well-studied Neyman--Pearson hypothesis test. This allows one to quantify network performance from a statistical risk perspective, which we consider to be an essential requirement for deployment in safety-critical applications. We therefore hope to contribute to broader research in safe and secure AI by aligning the field with traditional notions of risk in engineering.

{
\small
\bibliography{references}

@article{krizhevsky2009cifar,
  title={Cifar-10 and cifar-100 datasets},
  author={Krizhevsky, Alex and Nair, Vinod and Hinton, Geoffrey},
  journal={URl: https://www. cs. toronto. edu/kriz/cifar. html},
  volume={6},
  number={1},
  pages={1},
  year={2009}
}

@inproceedings{he2016deep,
  title={Deep residual learning for image recognition},
  author={He, Kaiming and Zhang, Xiangyu and Ren, Shaoqing and Sun, Jian},
  booktitle={Proceedings of the IEEE conference on computer vision and pattern recognition},
  pages={770--778},
  year={2016}
}

@article{barreno2007optimal,
  title={Optimal ROC curve for a combination of classifiers},
  author={Barreno, Marco and Cardenas, Alvaro and Tygar, J Doug},
  journal={Advances in Neural Information Processing Systems},
  volume={20},
  year={2007}
}

@article{anders1999model,
  title={Model selection in neural networks},
  author={Anders, Ulrich and Korn, Olaf},
  journal={Neural networks},
  volume={12},
  number={2},
  pages={309--323},
  year={1999},
  publisher={Elsevier}
}

@article{zhang2021survey,
  title={A survey on neural network interpretability},
  author={Zhang, Yu and Ti{\v{n}}o, Peter and Leonardis, Ale{\v{s}} and Tang, Ke},
  journal={IEEE transactions on emerging topics in computational intelligence},
  volume={5},
  number={5},
  pages={726--742},
  year={2021},
  publisher={IEEE}
}

@inproceedings{mandel2024permutation,
  title={Permutation-based hypothesis testing for neural networks},
  author={Mandel, Francesca and Barnett, Ian},
  booktitle={Proceedings of the AAAI Conference on Artificial Intelligence},
  volume={38},
  pages={14306--14314},
  year={2024}
}

@article{shwartz2017opening,
  title={Opening the black box of deep neural networks via information},
  author={Shwartz-Ziv, Ravid and Tishby, Naftali},
  journal={arXiv preprint arXiv:1703.00810},
  year={2017}
}

@article{saxe2019information,
  title={On the information bottleneck theory of deep learning},
  author={Saxe, Andrew M and Bansal, Yamini and Dapello, Joel and Advani, Madhu and Kolchinsky, Artemy and Tracey, Brendan D and Cox, David D},
  journal={Journal of Statistical Mechanics: Theory and Experiment},
  volume={2019},
  number={12},
  pages={124020},
  year={2019},
  publisher={IOP Publishing}
}

@article{kolchinsky2018caveats,
  title={Caveats for information bottleneck in deterministic scenarios},
  author={Kolchinsky, Artemy and Tracey, Brendan D and Van Kuyk, Steven},
  journal={arXiv preprint arXiv:1808.07593},
  year={2018}
}

@article{tax2017partial,
  title={The partial information decomposition of generative neural network models},
  author={Tax, Tycho MS and Mediano, Pedro AM and Shanahan, Murray},
  journal={Entropy},
  volume={19},
  number={9},
  pages={474},
  year={2017},
  publisher={MDPI}
}

@article{westphal2025generalized,
  title={A generalized information bottleneck theory of deep learning},
  author={Westphal, Charles and Hailes, Stephen and Musolesi, Mirco},
  journal={arXiv preprint arXiv:2509.26327},
  year={2025}
}

@article{montavon2019layer,
  title={Layer-wise relevance propagation: an overview},
  author={Montavon, Gr{\'e}goire and Binder, Alexander and Lapuschkin, Sebastian and Samek, Wojciech and M{\"u}ller, Klaus-Robert},
  journal={Explainable AI: interpreting, explaining and visualizing deep learning},
  pages={193--209},
  year={2019},
  publisher={Springer}
}

@inproceedings{ribeiro2016should,
  title={" Why should i trust you?" Explaining the predictions of any classifier},
  author={Ribeiro, Marco Tulio and Singh, Sameer and Guestrin, Carlos},
  booktitle={Proceedings of the 22nd ACM SIGKDD international conference on knowledge discovery and data mining},
  pages={1135--1144},
  year={2016}
}

@inproceedings{rawal2025evaluating,
  title={Evaluating Model Explanations without Ground Truth},
  author={Rawal, Kaivalya and Fu, Zihao and Delaney, Eoin and Russell, Chris},
  booktitle={Proceedings of the 2025 ACM Conference on Fairness, Accountability, and Transparency},
  pages={3400--3411},
  year={2025}
}

@article{wang2019nonparametric,
  title={Nonparametric density estimation for high-dimensional data—Algorithms and applications},
  author={Wang, Zhipeng and Scott, David W},
  journal={Wiley Interdisciplinary Reviews: Computational Statistics},
  volume={11},
  number={4},
  pages={e1461},
  year={2019},
  publisher={Wiley Online Library}
}

@article{kudithipudi2025neuromorphic,
  title={Neuromorphic computing at scale},
  author={Kudithipudi, Dhireesha and Schuman, Catherine and Vineyard, Craig M and Pandit, Tej and Merkel, Cory and Kubendran, Rajkumar and Aimone, James B and Orchard, Garrick and Mayr, Christian and Benosman, Ryad and others},
  journal={Nature},
  volume={637},
  number={8047},
  pages={801--812},
  year={2025},
  publisher={Nature Publishing Group UK London}
}

@article{eshraghian2023training,
  title={Training spiking neural networks using lessons from deep learning},
  author={Eshraghian, Jason K and Ward, Max and Neftci, Emre O and Wang, Xinxin and Lenz, Gregor and Dwivedi, Girish and Bennamoun, Mohammed and Jeong, Doo Seok and Lu, Wei D},
  journal={Proceedings of the IEEE},
  volume={111},
  number={9},
  pages={1016--1054},
  year={2023},
  publisher={IEEE}
}

@article{neyman1933ix,
  title={IX. On the problem of the most efficient tests of statistical hypotheses},
  author={Neyman, Jerzy and Pearson, Egon Sharpe},
  journal={Philosophical Transactions of the Royal Society of London. Series A, Containing Papers of a Mathematical or Physical Character},
  volume={231},
  number={694-706},
  pages={289--337},
  year={1933},
  publisher={The Royal Society London}
}

@misc{kingma2017adam,
      title={Adam: A Method for Stochastic Optimization}, 
      author={Diederik P. Kingma and Jimmy Ba},
      year={2017},
      eprint={1412.6980},
      archivePrefix={arXiv},
      primaryClass={cs.LG},
      url={https://arxiv.org/abs/1412.6980}, 
}

@article{paszke2019pytorch,
  title={Pytorch: An imperative style, high-performance deep learning library},
  author={Paszke, Adam and Gross, Sam and Massa, Francisco and Lerer, Adam and Bradbury, James and Chanan, Gregory and Killeen, Trevor and Lin, Zeming and Gimelshein, Natalia and Antiga, Luca and others},
  journal={Advances in neural information processing systems},
  volume={32},
  year={2019}
}

@article{lecun1998mnist,
  title={The MNIST database of handwritten digits},
  author={LeCun, Yann},
  journal={http://yann. lecun. com/exdb/mnist/},
  year={1998}
}

@inproceedings{perez2008kullback,
  title={Kullback-Leibler divergence estimation of continuous distributions},
  author={P{\'e}rez-Cruz, Fernando},
  booktitle={2008 IEEE international symposium on information theory},
  pages={1666--1670},
  year={2008},
  organization={IEEE}
}

@article{wang2009divergence,
  title={Divergence estimation for multidimensional densities via $ k $-Nearest-Neighbor distances},
  author={Wang, Qing and Kulkarni, Sanjeev R and Verd{\'u}, Sergio},
  journal={IEEE Transactions on Information Theory},
  volume={55},
  number={5},
  pages={2392--2405},
  year={2009},
  publisher={IEEE}
}

@article{zhao2020minimax,
  title={Minimax optimal estimation of KL divergence for continuous distributions},
  author={Zhao, Puning and Lai, Lifeng},
  journal={IEEE Transactions on Information Theory},
  volume={66},
  number={12},
  pages={7787--7811},
  year={2020},
  publisher={IEEE}
}

@inproceedings{kriener2022yin,
  title={The yin-yang dataset},
  author={Kriener, Laura and G{\"o}ltz, Julian and Petrovici, Mihai A},
  booktitle={Proceedings of the 2022 Annual Neuro-Inspired Computational Elements Conference},
  pages={107--111},
  year={2022}
}

@article{nielsen2011chernoff,
  title={Chernoff information of exponential families},
  author={Nielsen, Frank},
  journal={arXiv preprint arXiv:1102.2684},
  year={2011}
}

@article{chernoff1952measure,
  title={A measure of asymptotic efficiency for tests of a hypothesis based on the sum of observations},
  author={Chernoff, Herman},
  journal={The Annals of Mathematical Statistics},
  pages={493--507},
  year={1952},
  publisher={JSTOR}
}

@article{chernoff1956large,
  title={Large-sample theory: Parametric case},
  author={Chernoff, Herman},
  journal={The Annals of Mathematical Statistics},
  volume={27},
  number={1},
  pages={1--22},
  year={1956},
  publisher={JSTOR}
}

@misc{polyanskiy2016information,
  author       = {Yury Polyanskiy and Yihong Wu},
  title        = {Lecture Notes on Information Theory},
  year         = {2016},
  howpublished = {MIT OpenCourseWare},
  note         = {Course 6.441, Spring 2016},
  url          = {https://ocw.mit.edu/courses/6-441-information-theory-spring-2016/}
}

@article{kullback1951information,
  title={On information and sufficiency},
  author={Kullback, Solomon and Leibler, Richard A},
  journal={The annals of mathematical statistics},
  volume={22},
  number={1},
  pages={79--86},
  year={1951},
  publisher={JSTOR}
}

@article{zhu2021geometric,
  title={A geometric analysis of neural collapse with unconstrained features},
  author={Zhu, Zhihui and Ding, Tianyu and Zhou, Jinxin and Li, Xiao and You, Chong and Sulam, Jeremias and Qu, Qing},
  journal={Advances in Neural Information Processing Systems},
  volume={34},
  pages={29820--29834},
  year={2021}
}
\bibliographystyle{plainnat}
}





\newpage 

\appendix
\section{Binary Image Toy Dataset} \label{Binary Image Dataset Appendix}
We create a two-class, binary dataset with controllable parameters and tractable KL. Let $d \in \mathbb{N}$ be a dimension parameter and $p \in (0,1)$ a noise probability. Then our sample space is defined as:
\begin{equation}
\mathcal{X} = \{0,1\}^{d \times d}.
\end{equation}

We define two class-conditional distributions corresponding to the hypotheses $H_R: \text{row-white images}$ vs. $H_C: \text{column-white images}$ with equal priors. Under $H_R$, a row index is sampled uniformly in the dimension, $r \sim \mathcal{U}\{1, d\}$ and a template is generated as:
\begin{equation}
T^{(r)}_{ij} = \mathbf{1}\{ i = r \}
\end{equation}

that is, the $r$-th row is set to one and all remaining pixels are zero. The observed image $X$ is obtained by passing $T^{(R)}$ through an independent Binary Symmetric Channel with flip probability $p \in (0,1)$:

\begin{equation}
X_{ij} =
\begin{cases}
T^{(r)}_{ij} & \text{with probability } 1-p, \\
1 - T^{(r)}_{ij} & \text{with probability } p,
\end{cases}
\end{equation}

A similar process is followed for columns under $H_C \ (U^{(c)}_{ij} = \mathbf{1}\{ j = c \}$ for $c \sim \mathcal{U}\{1, d\}$). 

Note that the distribution of this dataset is supported on a finite alphabet and therefore the kNN KL estimator may not work well, especially for small $d$ due to distance ties and degenerate neighborhood volumes \cite{wang2009divergence}. However the same argument can be made for standard image datasets like MNIST and CIFAR-10 where pixel values are integer-valued (therefore distributed on a finite alphabet) but are normalized and treated as a compact subset of $[0,1]^d \subset \mathbb R^d$. Following this convention, we find empirically that for moderate $d$, the induced geometry is sufficient for stable estimation. We therefore fix $d=8$ (and $p=0.1$) as the canonical setting. 

We can derive the true KL divergence of this dataset as follows: first let $P_R$ and $P_C$ denote the row-white and column-white image conditionals on $\mathcal{X} = \{0,1\}^{d \times d}$. Then we have

\begin{equation}
D_{\mathrm{KL}} (P_R \| P_C) = \mathbb{E}_{X \sim P_R} \left[\log \frac{P_R(X)}{P_C(X)}\right]
\end{equation}

We can condition the marginals on the templates $t^r$ and $u^c$ representing the r-th row and c-th column white images (a.k.a uncorrupted) 
\begin{equation}
D_{\mathrm{KL}} (P_R \| P_C) = \mathbb{E}_{X \sim P_R} \left[\log \frac{\sum_{r=1}^d P(X | t^r)}{\sum_{c=1}^d P(X | c^r)}\right]
\end{equation}

since the uniform probability of picking $r$ or $c$ cancels out. Now for any template $s \in \{0,1\}^{d \times d}$, we can express the conditional likelihood under BSC noise for any image as: 
\begin{equation}
P(x \mid s) = \prod_{i=1}^d \prod_{j=1}^d
\begin{cases}
1-p & \text{if } x_{ij} = s_{ij}, \\
p   & \text{if } x_{ij} \neq s_{ij},
\end{cases}
\end{equation}

Denote the Hamming distance (pixels where they differ) between x and s as $d_H(x,s)$. Then we get the simplification 
\begin{equation}
P(x \mid s) = (1-p)^{d^2 - d_H(x,s)} p^{d_H(x,s)} = (1-p)^{d^2} (\frac{p}{1-p})^{d_H(x,s)} = (1-p)^{d^2} \lambda^{d_H(x,s)}
\end{equation}
where $\lambda = \frac{p}{1-p}$. Substituting this back into KL: 

\begin{equation}
D_{\mathrm{KL}} (P_R \| P_C) = \mathbb{E}_{X \sim P_R} \left[\log \frac{\sum_{r=1}^d \lambda^{d_H(x,t^r)} }{\sum_{c=1}^d \lambda^{d_H(x,u^c)}}\right]
\end{equation}

since the constants $(1-p)^{d^2}$ cancel. Now we need to derive expressions for the distance terms, starting with the row cases: 
\begin{align}
   d_H(x,t^r) &= \sum_{i=1}^d \sum_{j=1}^d \mathbf 1\{x_{ij} \neq t_{ij}^r\} = \sum_{j=1}^d \mathbf{1}\{ x_{rj} \neq t^{(r)}_{rj} \} + \sum_{\substack{i=1 \\ i \neq r}}^d \sum_{j=1}^d \mathbf{1}\{ x_{ij} \neq t^{(r)}_{ij} \} \\
           &= d - a_r(x) + T(x) - a_r(x) = d - 2a_r(x) + T(x)
\end{align}

where $T(x)$ is the sum over $x$ (meaning the number of ones in $x$) and $a_r(x)$ is the sum of the $r$-th row of $x$; the number of ones in row $r$ of $x$. We can do a similar calculation to get:
\begin{equation}
d_H(x,u^{(c)}) = d + T(x) - 2b_c(x)
\end{equation}

for the column templates with $b_c(x)$ as the number of ones in column $c$ of $x$. Plugging these distances back into the KL term, we observe that again $\lambda^{d + T(x)}$ cancels in each term:
\begin{equation}
D_{\mathrm{KL}} (P_R \| P_C) = \mathbb{E}_{X \sim P_R} \left[\log \frac{\sum_{r=1}^d \lambda^{- 2a_r(x)} }{\sum_{c=1}^d \lambda^{- 2b_c(x)}}\right] := \mathbb{E}_{X \sim P_R} [\phi(X)]
\end{equation}

by renaming the LLR as $\phi(X)$. Since $P_R$ is generated by uniformly sampling row values $r \sim U\{1, d\}$ (and the same for $P_C$) we can express the expectation as follows:
\begin{equation}
D_{\mathrm{KL}} (P_R \| P_C) = \frac{1}{d} \sum_{r=1}^d \mathbb{E}_{X \sim P(. | t^r)} [\phi(X)]
\end{equation}

Under a fixed template $t^r$, the pixels are distributed as independent Bernoulli variables, $X \sim P(. | t^r) $:
\begin{equation}
X_{ij} \sim \begin{cases}
    Bern(1-p) \ \ \text{if} \ i=r, \\
    Bern(p) \qquad \text{otherwise}
\end{cases}
\end{equation}

Therefore we can express the distributions $A_i = a_i(X)$ as binomials since they are the sums of pixel values distributed as i.i.d. Bernoulli variables. This gives us: 
\begin{equation}
A_i \sim Bin(d,p) \ \forall i \neq r \ \text{and} \ A_r \sim Bin(d,1-p)
\end{equation}

For the column variables, $B_j = b_j(X) = \sum_{i=1}^d X_{ij}$ we have to consider the pixel on the active row $i=r$ as a separate case since $X_{rj} \sim Bern(1-p)$ while $X_{ij} \sim Bern(p)$ for $i\neq r$. Then we get: 
\begin{equation}
B_j = Bern(1-p) + Bin(d-1, p)
\end{equation}
Plugging these distributions and taking the expectation of $\phi(X)$ (omitting subscript $X \sim P(. | t^r)$):
\begin{align}
    \mathbb{E}[\phi(X)] &= \mathbb{E}[\log \left(\sum_{i=1}^d \lambda^{-2A_i}\right)] - \mathbb{E}[\log \left(\sum_{j=1}^d \lambda^{-2B_j}\right)] \\
    &:= E_{row}(d,p) - E_{col}(d,p)
\end{align}

where $E_{row}$ and $E_{col}$ have closed form expressions in $(d,p)$ (omitted here for brevity) since $A_i$ and $B_j$ have distributions depending only in $d,p$ and are independent of the row choice $r$. Therefore the sum over rows cancels the $\frac{1}{d}$ factor outside and we are left simply with: 
\begin{equation}
D_{\mathrm{KL}} (P_R \| P_C) = E_{row}(d,p) - E_{col}(d,p) 
\end{equation}

\begin{figure}
    \includegraphics[width=0.3\linewidth]{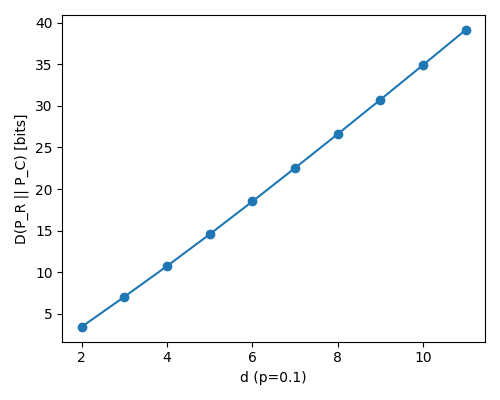}
    \includegraphics[width=0.3\linewidth]{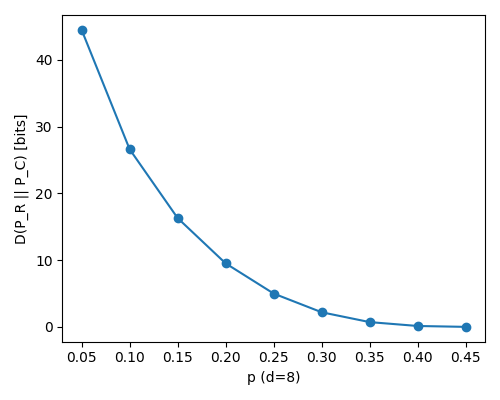}
    \includegraphics[width=0.4\linewidth]{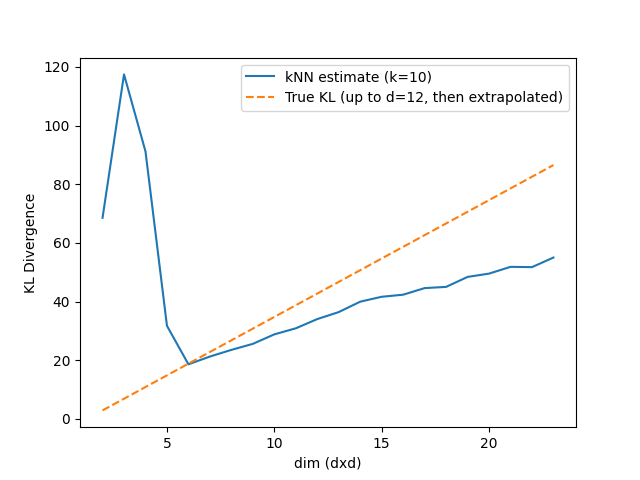}
    \caption{\textbf{KL Divergence of the Binary Image Dataset and its kNN Estimate:} We observe that for dimensions in the range 6-15, estimation is reliable but as $d$ grows, the downward bias becomes non-negligible. As discussed, for small $d$, the estimator likely explodes due to the small alphabet size of the underlying event space.}
    \label{fig:binary-image-dim-vs-kl}
\end{figure}

Figure \ref{fig:binary-image-dim-vs-kl} tracks the change of $D_{\mathrm{KL}} (P_R \| P_C)$ with $d$ and $p$ and compares it to the kNN estimator.

\section{Details and Proof of Theorem~\ref{thm:exact_kl}}
\label{app:kl_preservation}

\subsection{Derivation of the Cross-Entropy Gap}
\label{app:epsilon_kl}

Before stating the full proof, we first verify equation~\eqref{eq:epsilon_kl} from the main text. Writing
$p_{XY} = p_X p_{Y|X}$ and using the definition of $\epsilon(\theta)$:
\begin{align}
    \epsilon(\theta)
    &= -\mathbb{E}_{(X,Y)}\!\left[\log\frac{\sigma_Y(\theta(X))}{\eta_Y(X)}\right]
     = \mathbb{E}_{(X,Y)}\!\left[\log\frac{\eta_Y(X)}{\sigma_Y(\theta(X))}\right] \\
    &= \mathbb{E}_X\!\left[\mathbb{E}_{Y \sim \eta(\cdot|X)}\!\left[
        \log\frac{\eta_Y(X)}{\sigma_Y(\theta(X))}\right]\right] \\
    &= \mathbb{E}_X\!\left[\sum_{c=1}^K \eta_c(X)\,
        \log\frac{\eta_c(X)}{\sigma_c(\theta(X))}\right]
     = \mathbb{E}_X\!\left[D_{\mathrm{KL}}\!\left(\eta(X)\,\|\,
        \sigma(\theta(X))\right)\right].
\end{align}

\subsection{Assumptions}
Two assumptions are needed to show Theorem~\ref{thm:exact_kl}:

\begin{assumption}[Realizability]\label{ass:realizability}
    $\exists \, \tilde{\theta} \in \Theta$ such that $\sigma_c(\tilde{\theta}(x))
    = \eta_c(x)$ for all $c \in \mathcal{Y}$ and $\mu$-almost all $x \in
    \mathcal{X}$. That is, the function class is rich enough to recover the
    true posterior exactly. Since neural networks are universal approximators, in a very loose sense, this holds always but for a specific architecture and dataset, it is non trivial to verify.  
\end{assumption}

\begin{assumption}[Bounded Posteriors]\label{ass:bounded_posteriors}
    $\exists \, \kappa \in (0, 1/K)$ such that $\eta_c(x) \geq \kappa$ for
    all $c \in \mathcal{Y}$ and $\mu$-almost all $x \in \mathcal{X}$. This
    ensures $D_{\mathrm{inp}} < \infty$ and guarantees the LLR is well-defined
    and finite:
    \begin{align*}
        |\Lambda_c(x)|
        = \left|\log\frac{\eta_c(x)}{1 - \eta_c(x)}\right|
        \leq \log\frac{1-\kappa}{\kappa} < \infty.
    \end{align*}
    This is a requirement on the data itself and it prevents the degenerate case when perfect discrimination between all class-conditionals $(X_c, X_{c'}), \forall c, c' \in \mathcal Y$ is asymptotically possible. In practice we have ensured this by injecting a small amount of noise into the test samples before they get processed (see appendix \label{Infinite KL})
\end{assumption}

Now we are ready to state the proof.

\subsection{Proof}
\begin{proof}
    \textbf{Theorem~\ref{thm:exact_kl}}:
By Assumption~\ref{ass:realizability}, $\exists \, \tilde{\theta} \in \Theta$
with $\sigma_c(\tilde{\theta}(x)) = \eta_c(x)$ for all $c$ and $\mu$-a.e.\
$x$. Therefore:
\begin{align}
    \epsilon(\tilde{\theta})
    = \mathbb{E}_X\!\left[D_{\mathrm{KL}}\!\left(\eta(X) \,\|\,
        \sigma(\tilde{\theta}(X))\right)\right]
    = \mathbb{E}_X\!\left[D_{\mathrm{KL}}\!\left(\eta(X) \,\|\,
        \eta(X)\right)\right] = 0.
\end{align}
Since $\epsilon(\theta) \geq 0$ for all $\theta$ and $\theta^*$ minimizes
$\mathcal{L}(\theta)$; equivalently it minimizes $\epsilon(\theta)$, giving us
\begin{align}
    0 \leq \epsilon(\theta^*) \leq \epsilon(\tilde{\theta}) = 0,
\end{align}
so $\epsilon(\theta^*) = 0$. By the identity condition of KL divergence \citep{polyanskiy2016information},
$\sigma_c(\theta^*(x)) = \eta_c(x)$ for all $c \in \mathcal{Y}$ and
$p$-almost all $x$.


Similarly, since $\sigma_c(\theta^*(x)) = \eta_c(x)$ $\mu$-a.e., we have
\begin{align}
    \Lambda_c(x)
    = \log\frac{\eta_c(x)}{1-\eta_c(x)}
    = \log\frac{\sigma_c(\theta^*(x))}{1 - \sigma_c(\theta^*(x))}
    =: h_c(\theta^*(x)), \qquad \mu\text{-a.e.},
\end{align}
where $h_c$ is measurable and finite $\mu$-a.e.\ by
Assumption~\ref{ass:bounded_posteriors}. Therefore we have that $\theta^*(x)$ is a sufficient statistic for $(X_c, X_{\bar c})$

Indeed being sufficient statistic is a necessary and sufficient condition for KL divergence preservation \citep{polyanskiy2016information}. Therefore we have for probability measures $P \ll Q$ and measurable map $\theta$:
\begin{align}
    D_{\mathrm{KL}}(\theta(P) \| \theta(Q)) = D_{\mathrm{KL}}(P \| Q)
    \iff \theta(X) \text{ is a sufficient statistic for } (P, Q).
\end{align}

which gives immediately 
\begin{align}
    D_{\mathrm{KL}}(Z_c \| Z_{\bar{c}}) = D_{\mathrm{KL}}(X_c \| X_{\bar{c}}),
    \quad \forall c \in \mathcal{Y}.
\end{align}
Averaging over $c$:
\begin{align}
    D_{\theta^*}
    = \frac{1}{K}\sum_{c=1}^K D_{\mathrm{KL}}(Z_c \| Z_{\bar{c}})
    = \frac{1}{K}\sum_{c=1}^K D_{\mathrm{KL}}(X_c \| X_{\bar{c}})
    = D_{\mathrm{inp}},
\end{align}
hence $\delta(\theta^*) = 0$. 
\end{proof}

\subsection{Relaxing Realizability}
As stated, Realizability~\ref{ass:realizability} imposes a strict requirement on the network class $\Theta$ which may not hold or may not be easily verified. In that case, we conjecture that if one can further assume the network outputs are also bounded, there is a linear bound between $\delta(\theta)$ and $\epsilon(\theta)$. Formally defined:

\begin{assumption}[Bounded Model Output]\label{ass:bounded_model}
    $\exists \, \kappa' \in (0, 1/K)$ such that $\sigma_c(x) \geq \kappa'$ for
    all $c \in \mathcal{Y}$ and $\mu$-almost all $x \in \mathcal{X}$.
\end{assumption}

\begin{conjecture}[Approximate KL Preservation]
Under Assumptions~\ref{ass:bounded_posteriors} and~\ref{ass:bounded_model}, 
for any $\theta \in \Theta$:
\begin{align}
    \delta(\theta) \leq C(\kappa,\kappa') \, \epsilon(\theta),
\end{align}
In particular, $\delta(\theta) \to 0$ as $\epsilon(\theta) \to 0$, so that approximate 
cross-entropy minimization yields approximate KL preservation, with the rate controlled 
by a constant $C(\kappa,\kappa')$ depending on how bounded away from zero the true 
posteriors are.
\end{conjecture}
To show such a bound, first note that by the Donsker--Varadhan representation of KL 
divergence, we have:
\begin{align}
    D_{\mathrm{KL}}(X_c \| X_{\bar{c}})
    = \sup_{f_c \in L^1(X_{\bar{c}})}\left\{
        \mathbb{E}_{X_c}[f_c(X)]
        - \log\mathbb{E}_{X_{\bar{c}}}[e^{f_c(X)}]\right\}
\end{align}
where the supremum is attained by $f_c^* = \Lambda_c(x)$. Let $g_c(x) = 
\log\frac{\sigma_c(\theta(x))}{1-\sigma_c(\theta(x))}$ be the LLR of the model 
softmax, which is well defined under Assumption~\ref{ass:bounded_model}, and note:
\begin{align}
     D_{\mathrm{KL}}(X_c\|X_{\bar c}) \geq 
    \mathbb{E}_{X_c}[g_c] - \log\mathbb{E}_{X_{\bar c}}[e^{g_c}]
\end{align}
and therefore:
\begin{align}
    \delta(\theta) \leq D_{\mathrm{KL}}(X_c\|X_{\bar{c}}) - 
    \left[\mathbb{E}_{X_c}[g_c] - \log\mathbb{E}_{X_{\bar{c}}}[e^{g_c}]\right]
    = \mathbb{E}_{X_c}[\Lambda_c - g_c] + \log\frac{\mathbb{E}_{X_{\bar{c}}}[e^{g_c}]}{\mathbb{E}_{X_{\bar{c}}}[e^{\Lambda_c}]}
\end{align}
Both terms on the right hand side vanish when $\epsilon(\theta) = 0$, since  Theorem~\ref{thm:exact_kl} gives $\sigma_c(\theta(x)) = \eta_c(x)$ $\mu$-a.e., 
and hence $g_c = \Lambda_c$ $\mu$-a.e. 

The key observation is that  $\Lambda_c - g_c$ involves the same log ratios as $\epsilon(\theta)$, but the  expectation is taken under $p(x|Y=c)$ rather than $p(x)$. Establishing a linear bound therefore reduces to controlling this change of measure where we presume the constant factor $C(\kappa,\kappa')$ will naturally arise.

\section{kNN Estimation of KL Divergence}\label{app:kNN-KL}

\subsection{Finite Sample Bias} \label{kNN KL Bias}
The proposed kNN estimators in \cite{perez2008kullback} and \cite{wang2009divergence}, which we extensively utilize in our experiments, are known to be consistent for any value of $k$, asymptotically. However with finite samples, they are biased; this bias typically scales with dimension. \cite{wang2009divergence} suggests using adaptive $k$ values pointwise for good log-density ratio estimation, which is proven to achieve optimal bias/variance trade-offs but is also computationally expensive. 

To overcome the finite sample bias with negligible compute overhead, we use the following simple heuristic for selecting $k$ in a data dependent way. (Note that we don't claim any guarantees or bias bounds as this depends on the geometry of the underlying distribution):

\begin{algorithm}[tbh]
\caption{Null-consistency based selection of $k$ for $k$NN divergence estimation}
\label{alg:knn_null_cons}
\begin{algorithmic}[1]
\Require Dataset $\mathcal{X} = \{x_i\}_{i=1}^N$, candidate set $\mathcal{K}$, number of splits $S$
\State Initialize $k^* \gets 1$

\For{$k \in \mathcal{K}$}
    \State $b_k \gets 0$
    
    \For{$s = 1, \dots, S$}
        \State Sample a random partition $\mathcal{X} = \mathcal{X}_1^{(s)} \cup \mathcal{X}_2^{(s)}$, with $\mathcal{X}_1^{(s)} \cap \mathcal{X}_2^{(s)} = \emptyset$
        \State $\widehat{D}_k^{(s)} \gets \widehat{D}_k\!\left(\mathcal{X}_1^{(s)} \,\|\, \mathcal{X}_2^{(s)}\right)$
        \State $b_k \gets b_k + \widehat{D}_k^{(s)}$
    \EndFor
    
    \State $b_k \gets b_k / S$
\EndFor

\State $k^* \gets \arg\min_{k \in \mathcal{K}} |b_k|$
\State \Return $k^*$
\end{algorithmic}
\end{algorithm}
Since the true divergence of any distribution $X$ with itself $D_{\mathrm{KL}} (X \| X) = 0$, this procedure outputs the $k$ that is most stable in low divergence regimes. 

Furthermore \cite{zhao2020minimax} notes that when the data lives on a lower-dimensional manifold than its original dimension, the kNN estimator has upward bias as the neighborhoods grow disproportionately. This bias again scales exponentially with dimension. Since neural network logits of a classifier with $K$ outputs is known to collapse to a $K-1$ dimensional manifold (so-called neural collapse \cite{zhu2021geometric}), we apply a change of coordinate before network outputs are fed into the KL estimator. Formally: 

Let $Z=(Z_1,\dots,Z_K)^\top\in\mathbb{R}^K$ be the logit vector and $\sigma(Z)$ denote the softmax vector. Then we have $\sigma(Z) = sigma(Z + c\mathbf{1})$ for any constant $c \in \mathbb{R}$. Without loss of generality, (as it can be defined for any reference class) define the following projection $\Pi \in \mathbb{R}^{K \times K-1}$: 
\begin{equation}
u = \Pi Z = (Z_1-Z_K,\dots,Z_{K-1}-Z_K)^\top \in \mathbb{R}^{K-1}.
\end{equation}
Then $\Pi(Z+K\mathbf 1)=\Pi Z$ and 
\begin{equation}
\sigma(Z)_i = \frac{e^{u_i}}{1 + \sum_{j=1}^{K-1} e^{u_j}}
\end{equation} 
with $u_K := 0$ for completeness. So $u$ is a smaller, shift-invariant representation of the logits and we can estimate KL divergence on samples of $u$ instead of $Z$.

\subsection{Singularities in KL-Divergence}\label{Infinite KL}
Recall that the KL divergence between $P$ and $Q$ can be infinite, in the case  when $P$ is not \textit{absolutely continuous} with respect to $Q$, i.e.,
\begin{equation}
\exists\, A \subset \mathcal X \text{ measurable such that } P(A) > 0 \ \text{and}\ Q(A) = 0.
\end{equation}

Far from being an edge case, this can easily happen when we define $P$ and $Q$ as the class conditioned distributions; when $y = f(x)$ for a pair $(x,y)$ and deterministic $f$, we immediately have $P(x,y) > 0$ while $Q(x,y) = 0$; in fact the distributions $X_c$ and $X_{c'}$ may have completely disjoint support for class pairs $c, c'$.

To ensure absolute continuity between class conditionals, we inject a small amount of white noise to each input sample before it is fed into the divergence estimator. Note that unless the corrupted samples are fed into the network, DPI does not hold. Formally let $\tilde{X_{c}^{\sigma}} = X_c + \mathcal{N}(0, \sigma^2)$ and $\tilde{Z}_c = \theta (\tilde{X_{c}^{\sigma}})$ with $\theta$, the network. Then, we \textit{only} have the chain:
\begin{equation}
D_{\mathrm{KL}}(X_c \| X_{\bar{c}}) > D_{\mathrm{KL}}(\tilde{X_c}^\sigma \| \tilde{X}_{\bar{c}}^\sigma) \geq D_{\mathrm{KL}}(\tilde{Z}_c \| \tilde{Z}_{\bar{c}})
\end{equation}

regardless of whether $X_c \ll X_{\bar{c}}$ or $X_c \nll X_{\bar{c}}$. In the small noise limit we recover 
\begin{equation}
\lim_{\sigma \to 0} \ D_{\mathrm{KL}}(\tilde{X_c}^\sigma \| \tilde{X}_{\bar{c}}^\sigma) = D_{\mathrm{KL}}(X_c \| X_{\bar{c}}) 
\end{equation}

\begin{figure}[h]
    \centering
    \includegraphics[width=0.3\linewidth]{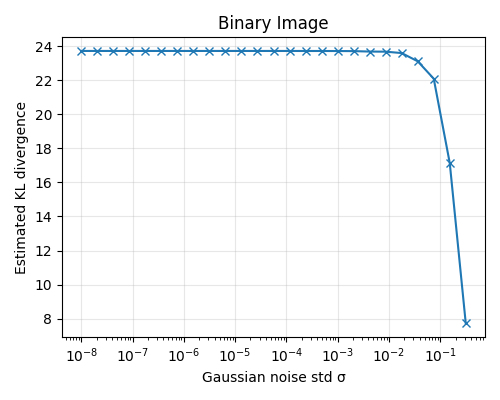}
    \includegraphics[width=0.3\linewidth]{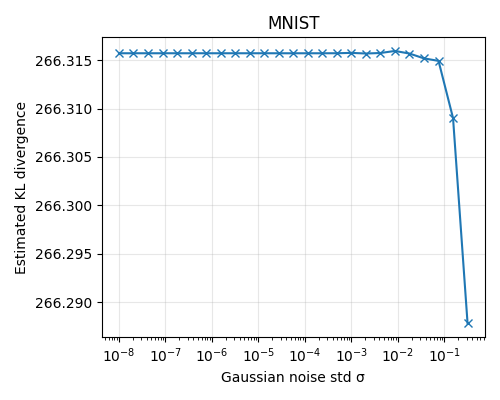}
    \includegraphics[width=0.3\linewidth]{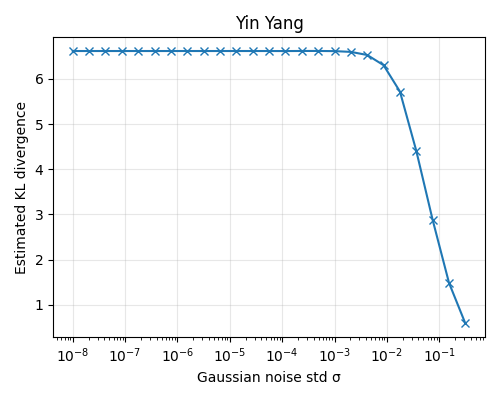}
    \caption{\textbf{Effect of additive noise on KL Div estimates}: Note that for noise values below $10^{-3}$, the estimates are stable.}
    \label{fig:noise-vs-kl}
\end{figure}

To observe the effect of this noise on the estimators (see Figure \ref{fig:noise-vs-kl}), we sweep across a logarithmic range of additive noise values and track the estimator output
$
D_\sigma(X) = \frac{1}{K} \sum_{c \in Y} \hat{D}(\tilde{X_{c}^{\sigma}} \| \tilde{X}_{\bar{c}}^\sigma)
$ We observe that for noise values with roughly $\sigma < 10^{-3}$, the added noise has negligible effect on the estimated KL, while ensuring theoretical consistency. Therefore for all our experiments we inject noise with $\sigma = 10^{-6}$.

\section{Experimental Details}
\subsection{SNN Membrane Potential Distribution Support Size} \label{SNN Finite Alphabet}
Recall the Leaky-Integrate Fire neuron dynamics: 
\begin{equation}
U_l[t] = \eta U_l[t-1] + W_{l-1} S_{l-1}[t] - V_{\mathrm{th}}S_l[t-1], \hspace{0.7cm} 
    S_l[t] = \Theta\!\left(U_l[t] - V_{\mathrm{th}}\right)
\end{equation} and note that $t \in \{1, 2 ... \tau\}$ for $\tau \in \mathbb{N}$ constant (in our experiments $\tau = 5$). Without loss of generality, assume a single neuron for each layer to simplify computation. Since each spike event is binary, the number of possible values of $U_l$ given a fixed weight $W_{l-1}$ is constant. Observe that each update adds one of 4 values: 
\begin{equation}
\Delta U_l[t]\in\{0,\; W_{l-1},\; -V_{\mathrm{th}},\; W_{l-1}-V_{\mathrm{th}}\}.
\end{equation} and since $U_l[0] = 0$, the unrolled $U_l[t]$ is: 
\begin{equation}
U_l[t] = \sum_{k=1}^{t} \eta^{t-k}
\bigl(W_{l-1} S_{l-1}[k] - V_{\mathrm{th}} S_l[k-1] \bigr)
\end{equation}
Therefore 
\begin{equation}
\bigl|\operatorname{supp}(U_l[t])\bigr| \le 4^{t} \ \rightarrow \ \bigl|\operatorname{supp}(\{U_l[t]\}_{t=1}^{\tau})\bigr|
\le \sum_{t=1}^{\tau} 4^{t}
= \frac{4^{\tau+1}-4}{3}
\end{equation}

\subsection{Counter Example for Majority Voting} \label{Maj vote counterex}
Consider a binary classifier $\theta$ and 3 samples: $(x_1, x_2, x_3) \sim P^{\otimes 3}$ from the same class with the following probability of error
\begin{equation}
p := \mathbb{P}[\theta(x) \neq Y | Y]
\end{equation}
Let now $\theta_{maj}$ be the majority classifier using $\theta$ on 3 samples. Then the error probability of $\theta_{maj}$ is:
\begin{align}
    p_{maj}:= \mathbb{P}[\theta_{maj}(x_1, x_2, x_3) \neq Y | Y] =& \mathbb{P}[\text{at least 2 samples are classified wrong}] \\
    =& {{3}\choose{2}} p^2(1-p) + p^3 = 3p^2 - 2p^3
\end{align}

Then for any $p \in (\frac{1}{2}, 1)$, we have that $p_{maj} > p$; though arguably this is not very realistic as $p \in (\frac{1}{2}, 1)$ corresponds to worse than random classification.




\end{document}